\documentclass[preprint,12pt,authoryear]{elsarticle}

\usepackage{algorithm}
\usepackage{mdframed}
\usepackage{float}
\usepackage{tabularx}
\usepackage{amsmath}
\usepackage{changepage}
\usepackage{subcaption}
\usepackage{tcolorbox}
\usepackage[noend]{algpseudocode}
\usepackage{float}
\usepackage{amssymb}
\usepackage{multirow}
\usepackage{amssymb}
\usepackage[labelfont={color=black}]{caption}
\usepackage{xcolor}
\usepackage{hyperref}
\hypersetup{
	colorlinks   = true,
	citecolor    = blue,
	linkcolor = blue  
}
\usepackage{xstring}
\makeatletter
\AtBeginDocument{
	\let\oldref\ref
	\renewcommand{\ref}[1]{
		\IfBeginWith{#1}{fig:}%
		{{\color{blue}Figure~\oldref{#1}}}%
		\IfBeginWith{#1}{eqn:}%
		{{\color{blue}\oldref{#1}}}%
		{\IfBeginWith{#1}{tab:}{{\color{blue}Table~\oldref{#1}}}{Unsupported ref start}}}%
}
\makeatother

\newcommand\norm[1]{\left\lVert#1\right\rVert}

\journal{CSSC. Accepted on Apr 28, 2020.}

\begin{document}

\begin{frontmatter}

\title{Beyond the EM Algorithm: Constrained Optimization Methods for Latent Class Model}

\author{Hao Chen\fnref{label5}}
\ead{hao.chen@stat.ubc.ca}
\fntext[label5]{Corresponding Author. We thank the editor and the anonymous reviewer for suggestions that improved the manuscript.}
\address{Research \& Development, Precima, Chicago, IL USA 60631}

\author{Lanshan Han}
\ead{lhan2@precima.com}
\address{Research \& Development, Precima, Chicago, IL USA 60631}

\author{Alvin Lim}
\ead{alim@precima.com}
\address{Research \& Development, Precima, Chicago, IL USA 60631}

\begin{abstract} \label{sec:abstract}
Latent class model (LCM), which is a finite mixture of different categorical distributions, is one of the most widely used models in statistics and machine learning fields. Because of its non-continuous nature and flexibility in shape, researchers in areas such as marketing and social sciences also frequently use LCM to gain insights from their data. One likelihood-based method, the Expectation-Maximization (EM) algorithm, is often used to obtain the model estimators. However, the EM algorithm is well-known for its notoriously slow convergence. In this research, we explore alternative likelihood-based methods that can potential remedy the slow convergence of the EM algorithm. More specifically, we regard likelihood-based approach as a constrained nonlinear optimization problem, and apply quasi-Newton type methods to solve them. We examine two different constrained optimization methods to maximize the log-likelihood function. We present simulation study results to show that the proposed methods not only converge in less iterations than the EM algorithm but also produce more accurate model estimators.

(NOTE: the paper has been published online at \href{https://doi.org/10.1080/03610918.2020.1764034}{here}) 
\end{abstract}


\end{frontmatter}


\section{Introduction} \label{sec:intro}
Latent class model (LCM) (\cite{mccutcheon1987latent}) is a model to study latent (unobserved) categorical variables by examining a group of observed categorical variables which are regarded as the indictors of the underlying latent variables. It can be regarded as a special case of the finite mixture model (FMM) with component distributions being categorical distributions. It is widely used to analyze ordered survey data collected from real world applications. In many applications in econometrics, social sciences, biometrics, and business analytics (see \cite{Hagenaars2002latent,oser2013online} for example), finite mixture of categorical distributions arises naturally when we sample from a population with heterogeneous subgroups. LCM is a powerful tool to conduct statistical inference from the collected data in such situations. \\

We provide a motivating example from \cite{white2014bayeslca} where an LCM is applied to analyze a dataset of patient symptoms recorded in the Mercer Institute of St. James' Hospital in Dublin, Ireland (\cite{moran2004syndromes}). The data is a recording of the presence of six symptoms displayed by $240$ patients diagnosed with early onset Alzheimer's disease. The six symptoms are as follows: hallucination, activity, aggression, agitation, diurnal and affective, and each symptom has two states: either present or absent. \cite{white2014bayeslca} proposed to divide patients into $ K =3 $ groups such that patients are homogeneous within each group and heterogeneous between groups. Each group's characteristics are summarized by the LCM parameters that help doctors prepare more specialized treatments. In this sense, LCM is a typical unsupervised statistical learning method that could ``learn" the group labels based on the estimated parameters. \\

Due to its theoretical importance and practical relevance, many different approaches have been proposed to estimate the unknown parameters in LCMs from the observed data. In general, there are mainly two different paradigms. The first one is the frequentist's approach of maximum likelihood estimation (MLE), i.e., one maximizes the log-likelihood as a function of the unknown parameters. In contrast, a second paradigm -- the Bayesian approach -- where the unknown parameters obey a distribution and assumes prior distributions on them, then one either analytically or numerically obtains the posterior distributions and statistical inference is carried out based on the posterior distributions. \\

In recent years, significant progress has been made on Bayesian inference in LCM. \cite{white2014bayeslca}, by assuming the Dirichlet distribution on each unknown parameter, used Gibbs sampling to iteratively draw samples from the posterior distribution and then conduct inference on the LCM using the samples drawn. The authors also provided an implementation of the approach in R. \cite{li2018bayesian} described a similar Bayesian approach to estimate the parameters and they also utilized the Dirichlet distribution as the prior distribution.  \cite{asparouhov2011using} introduced a similar implementation package of Bayesian LCM in Mplus. However, compared to the fast development of the Bayesian inference via Markov chain Monte Carlo (MCMC), the frequentist's MLE approach for LCM has largely lagged. As far as we know, researchers still heavily rely on the expectation-maximization (EM) algorithm (\cite{dempster1977maximum}), even with its notoriously slow convergence (see for instance \cite{meilijson1989fast}), to maximize the log-likelihood function. It is known that some authors (\cite{jamshidian1997acceleration}) use Quasi-Newton methods as alternatives for the EM algorithm in Gaussian mixture models. However, the extension to LCM is not straightforward since LCM includes a lot more intrinsic constraints on the parameters than the general Gaussian mixture model when considered as an optimization problem. More sophisticated optimization methods need to be applied when maximizing the log-likelihood function.\\

This paper primarily focuses on the MLE paradigm. We propose the use of two widely-used constrained optimization methods to maximize the likelihood function, namely, the Projected Quasi-Newton method and the sequential quadratic programming method. Our contributions include not only exploring alternatives beyond the EM algorithm, but also demonstrating that better results could be obtained by using these alternatives. The rest of this paper is organized as follows: in Section~\oldref{sec:loglikelihood}, we present the preliminaries including the log-likelihood function and the classical EM algorithm. In Section~\oldref{sec:opt}, we introduce and discuss the two constrained optimization methods in detail. Some simulation studies and a real world data analysis are presented in Section~\oldref{sec:sim} to compare the performance of the proposed methods with the EM algorithm. We make concluding remarks in Section~\oldref{sec:com}.

\section{Latent Class Models and the EM Algorithm}  \label{sec:loglikelihood}
In many applications, a finite mixture distribution arises naturally when we sample from a population with heterogeneous subgroups, indexed by $k$ taking values in $\{1,\cdots, K\}$. Consider a population composed of $K$ subgroups, mixed at random in proportion to the relative group sizes $\eta_1,\cdots,\eta_K$. There is a random feature $y$, heterogeneous across and homogeneous within the subgroups.
The feature $y$ obeys a different probability distribution, often from the same parametric family $p(y|\theta)$ with $\theta$ differing, for each subgroup. Now we sample from this population, if it is impossible to record the subgroup label, denoted by $s$, then the density $p(y)$ is:
$$p(y) \, =\, \sum_{k=1}^K \eta_k p(y|\theta_k), $$
which is a finite mixture distribution. In this situation, we often need to estimate the $\theta_k$'s as well as $\eta_k$ based on the random samples of $y$, when the subgroup label $s$ is known or unknown. Throughout this paper, we assume that $K$ is known.

The LCM is a special case of the FMM. In LCM, the component densities are multivariate categorical distributions. That is, $\boldsymbol{y}=(\boldsymbol{y}_1,\cdots,\boldsymbol{y}_d)$ with each $\boldsymbol{y}_j$ being a categorical random variable, taking values from $c_j$ categories $\{1,\cdots,c_j\}$. It is assumed that $\boldsymbol{y}_j$'s are independent within each subgroup with an indictor $s$ (the latent variable), which is a categorical random variable taking values in $\{1,\cdots,K\}$, i.e., within each subgroup, the probability density function (PDF) is written as:
$$\prod_{j=1}^d \prod_{l=1}^{c_j} \pi_{k,j,l}^{\mathcal I(\boldsymbol{y}_{j}=l )},$$
where $\pi_{k,j,l} = \mbox{Pr}(\boldsymbol{y}_j=l|s=k)$ and $\mathcal I(\cdot)$ is the Iverson bracket function, i.e.
$$\mathcal I(\mbox{P}) \, = \, \left\{
\begin{array}{ll}
1 & \mbox{if P is true;}\\
0 & \mbox{if P is false.}
\end{array}
\right.$$
Overall, the mixture density of latent class models is:
$$p(\boldsymbol{y}|\boldsymbol{\theta}) \, = \, \sum_{k=1}^K \left(\eta_k \prod_{j=1}^d \prod_{l=1}^{c_j} \pi_{k,j,l}^{\mathcal I(\boldsymbol{y}_j=l)}\right),$$
where, the parameters $\boldsymbol{\theta}$ include both the weight distribution $\eta$ and the $\pi_{k,j,l}$'s that define the categorical distributions.

Suppose we have collected $ N $ samples drawn from the LCM distribution, denoted by $ \{ y^1, \cdots, y^N \} $. We write $ Y = \left[y^1, \cdots, y^N\right]^{T} \in \mathbb R^{N \times d}$ as the data matrix. The log-likelihood function is given by
\begin{eqnarray} \label{eqn:loglikelihood}
L(\boldsymbol{\theta}| Y) &=& \log\left( \prod_{i=1}^{N} p(\boldsymbol{y}^i | \boldsymbol{\theta}) \right) \nonumber \\
&=& \sum_{i=1}^{N} \log\left( \sum_{k=1}^K \eta_k  \prod_{j=1}^{d} \prod_{l = 1}^{c_d} \pi_{k, j, l}^{\mathcal I(y^i_j=l)}\right).
\end{eqnarray}
The maximum likelihood principle is to find a $\boldsymbol{\theta}^*$ that maximizes the log-likelihood function (\oldref{eqn:loglikelihood}) as the estimation of $\boldsymbol{\theta}$. Clearly, we can regard the problem of finding such a $\boldsymbol{\theta}^*$ as an optimization problem. At the same time, we notice that the LCM implies several constraints that need to be satisfied when maximizing the log-likelihood function (\oldref{eqn:loglikelihood}). In particular, the $\eta_k$'s are all nonnegative and sum up to 1. Also, for each $k=1,\cdots,K$ and $j=1,\cdots,d$, the $\pi_{k,j,l}$'s are all nonnegative and sum up to 1. Let $\eta = (\eta_k)_{k=1}^K$ be the vector of $\eta_k$'s and $\pi = \left(\pi_{k,j,l}\right)_{k=1,\cdots,K;j=1,\cdots,d;l=1,\cdots,c_j}$ be the vector of $\pi_{k,j,l}$'s. From an optimization point of view, the MLE in the LCM case is the following optimization problem.
\begin{equation} \label{eqn:LMEasOptimization}
\begin{array}{rl}
\displaystyle{\max_{\eta, \pi} }& \displaystyle{\sum_{i=1}^{N} \log\left( \sum_{k=1}^K \eta_k  \prod_{j=1}^{d} \prod_{l = 1}^{c_d} \pi_{k, j, l}^{\mathcal I(y^i_j=l)}\right) } \\[5pt]
\mbox{s.t.} & \displaystyle{\sum_{k=1}^K \eta_k } \, = \, 1,\\[5pt]
&\displaystyle{\sum_{l=1}^{c_j} \pi_{k, j, l} }\, = \, 1, \, \forall \, k=1,\cdots,K, j=1,\cdots,d,\\
&\eta_k \, \geq \, 0, \forall \, k=1,\cdots,K, \\
&\pi_{k,j,l} \,  \geq \, 0, \forall \, k=1,\cdots,K, j=1,\cdots, d, l =1,\cdots,c_j.
\end{array}
\end{equation}
As we can see, the optimization problem (\oldref{eqn:LMEasOptimization}) possesses $ K \times d + 1 $ equality constraints together with nonnegativity constraints on all the individual decision variables. While there are considerable number of constraints, the feasible region in (\oldref{eqn:LMEasOptimization}) is indeed the Cartesian product of $K\times d + 1$ probability simplexes. We recall that a probability simplex in $n$-dimensional space $\mathbb R^n$ is defined as
$$ \mathcal P^{n} \, = \, \left\{x\in \mathbb R^n_+ \left| \sum_{i=1}^n x_i = 1\right. \right \},$$
where $\mathbb R^n_+$ is the nonnegative orthant of $\mathbb R^n$. Let $\pi_{k,j} = \left(\pi_{k,j,l}\right)_{l=1}^{c_j}$ for all $k=1,\cdots,K$ and $j=1,\cdots,d$. The constraints in (\oldref{eqn:LMEasOptimization}) can be written as $\eta \in \mathcal P^K$ and $\pi_{k,j} \in \mathcal P^{c_j}, \,\, \forall \, k=1,\cdots,K;j=1,\cdots,d$.\\

To maximize the log-likelihood function in (\oldref{eqn:loglikelihood}), the EM algorithm is a classical approach.  In statistics, the EM algorithm is a generic framework that is commonly used in obtaining maximum likelihood estimators. The reason why the EM algorithm enjoys its popularity in finite mixture model is the fact that we can view finite mixture model as an estimation problem with missing data. More specifically, if we know the true label of each observation, we could obtain the MLE in a fairly straightforward fashion. On the other hand, if we know the true model parameters, it is also trivial to compute the probability each observation belonging to each class. Therefore, a natural idea is that we begin the process with an initial random guess of the parameters, and compute the probability each observation belonging to each class E(xpectation)-step. With those probabilities we compute the MLE, which is the M(aximization)-step. We iterate between the two steps until a convergence condition is reached. Particularly for the LCM, when the EM algorithm is applied to it, the constraints are implicitly satisfied for all the iterations thanks to the way the EM algorithm updates the values of the parameters. This nice property does not necessarily hold naturally when other non-linear optimization algorithms are applied to the optimization problem (\oldref{eqn:LMEasOptimization}).  \\

In the context of LCM, the details of the EM algorithm is given in Algorithm \oldref{alg:EM}. We make two comments on Algorithm \oldref{alg:EM}. First, Algorithm \oldref{alg:EM} does not produce standard errors of MLE as a by-product. In order to conduct statistical inference, one has to compute the observed Fisher information matrix and it could be algebraically tedious or might only apply to special cases. This is one of the criticisms often laid out against the EM algorithm as compared to Bayesian analysis using Gibbs samplers for example, where independent posterior samples are collected and statistical inference is easy under such circumstance. Second, the convergence of Algorithm \oldref{alg:EM} is typically slow. \cite{wu1983convergence} studied the convergence issue of the EM algorithm and concluded that the convergence of the EM algorithm is sublinear when the Jacobian matrix of the unknown parameters is singular. \cite{jamshidian1997acceleration} also reported that the EM algorithm could well be accelerated by the Quasi-Newton method. In Section~\oldref{sec:sim}, we shall also empirically observe the two constrained optimization methods converge in less iterations than the EM algorithm.
\begin{algorithm}[htbp!]
	\caption{EM Algorithm for Latent Class Model}\label{alg:EM}
	\begin{algorithmic}[1]
		\State Supply an initial guess of the parameters $$\boldsymbol{\theta}^{(0)} = (\eta_k^{(0)},\pi_{j,k,l}^{(0)})_{k=1,\cdots,K; j=1,\cdots,d; l=1,\cdots,c_j}$$ and a convergence tolerance $\epsilon$.
		\State Initialize with $t = 1 $, and $\Delta > \epsilon$.
		\State \textbf{While} $ \Delta > \epsilon$:
        \State \indent (E-step) For each $i=1,\cdots,N$ and $k=1,\cdots,K$, compute:
        $$D_{ik}^{(t)} \, = \, \frac{\eta_k^{(t-1)} \prod_{j=1}^d \prod_{l=1}^{c_j} \left(\pi^{(t-1)}_{k,j,l}\right)^{\mathcal I(y^i_j=l)}}{\sum_{k=1}^K \eta_k^{(t-1)} \prod_{j=1}^d \prod_{l=1}^{c_j} \left(\pi^{(t-1)}_{k,j,l}\right)^{\mathcal I(y^i_j=l)}}.$$
		\State  \indent (M-step for weights) Compute: $$ \hat{\eta}_k^{(t)} = \frac{1}{N} \sum_{i=1}^{N} D_{ik}^{(t)}. $$
		\State  \indent (M-step for categorical parameters) Compute: $$  \pi^{(t)}_{k,j,l} \, = \, \frac{ \sum_{i=1}^n D_{ik}^{(t)} \left[\sum_{j=1}^d \sum_{l=1}^{c_j} \mathcal I(y_j^i=l) \right] }{\sum_{l=1}^{c_j} \left\{ \sum_{i=1}^n D_{ik}^{(t)} \left[\sum_{j=1}^m \sum_{l=1}^{c_j} \mathcal I(y_j^i=l) \right] \right\}}. $$
		\State  \indent Compute $\Delta = \norm{\boldsymbol{\theta}^{(h+1)} - \boldsymbol{\theta}^{(h)}}_1.$
		\State  \indent $t = t + 1.$
	\end{algorithmic}
\end{algorithm}

\section{Constrained Optimization Methods}  \label{sec:opt}

Motivated by the significant progress in constrained non-linear optimization, as well as the constrained nature of the LCM estimation problem, we propose to apply two non-linear optimization approaches to solve the optimization problem (\oldref{eqn:LMEasOptimization}). We notice that the EM algorithm is closely related to a gradient decent method \cite{wu1983convergence}, whose convergence rate is at most linear. On the other hand, it is known in optimization theory that if the second order information is utilized in the algorithm, quadratic convergence may be achieved, e.g., the classical Newton's method. However, in many applications, it is often computationally very expensive to obtain the second order information, i.e., the Hessian matrix. One remedy is to use computationally cheap approximation of the Hessian matrix. This idea leads to the family of Quasi-Newton methods in the unconstrained case. While the convergence rate is typically only superlinear, the per iteration cost (both the execution time and the memory usage) is significantly reduced. In the constrained case, sophisticated methods have been developed to allow us to deal with the constraints. Given that it is relative easy to solve a constrained optimization problem when the objective function is quadratic and the constraints are all linear, one idea in constrained non-linear optimization is to approximate the objective function (or the Lagrangian function) by a quadratic function (via second-order Taylor expansion at the current solution) and approximate the constraints by linear constraints (via first-order Taylor expansion at the current solution). A new solution is obtained by solving the approximation and hence a new approximation can be constructed at the new solution. Analogous to the idea of quasi-Newton methods in the unconstrained case, in the constrained case, we can also consider an approximated Taylor expansion without having to compute the Hessian matrix exactly. Once an approximated quadratic program is obtained, one may use different approaches to solve it. For example, one can use an active set method or an interior point method to solve the quadratic program when it does not possess any specific structure. When the feasible region of the quadratic program is easily computable (typically in strongly polynomial time), a gradient projection method can be applied to solve the quadratic program approximation. As we have seen, the feasible region of optimization problem (\oldref{eqn:LMEasOptimization}) is the Cartesian product of probability simplexes. It is known that projection on a probability simplex is computable in strongly polynomial time. Therefore, it is reasonable to apply a projection method to solve the quadratic program approximation. In the following subsections, the two approaches we propose are discussed in details. In both approaches, we need to evaluate the gradient of the LCM log-likelihood function. We provide the analytical expression below.
For the $\eta$ part, we have:
\begin{equation}\label{eqn:likelihood_gradient_eta}
\frac{\partial  L}{\partial \eta_k} \, =\, \sum_{i=1}^n \left[ \frac{f(y^i|\theta_k)}{\left(\sum_{k=1}^K \eta_k f(y^i|\theta_k) \right) } \right], \,\, k=1,\cdots,K.
\end{equation}
For the $\pi$ part, we have for all $i=1,\cdots,n;k=1,\cdots,K;j=1,\cdots,m; l=1,\cdots,c_j$:
\begin{equation}\nonumber
\frac{\partial f(y^i|\pi_k)}{\partial \pi_{k,j,l}} \,  = \, \mathcal I(y^i_j=l) \prod_{\jmath \neq j}\prod_{\ell=1}^{c_j} \pi_{k,\jmath,\ell}^{\mathcal I(y^i_\jmath=\ell)},
\end{equation}
where $\pi_k = (\pi_{k,j,l})_{j=1,\cdots,d;l=1,\cdots,c_j}$.
And therefore, for all $k=1,\cdots,K$,
\begin{equation}\label{eqn:likelihood_gradient_pi}
\frac{\partial L}{\partial \pi_k} \, =\, \sum_{i=1}^n \left[ \frac{\eta_k}{\left(\sum_{k=1}^K \eta_k f(y^i|\pi_k) \right) } \frac{\partial f(y^i|\pi_k)}{\partial \pi_k}\right].
\end{equation}
\subsection{Limited Memory Projected Quasi-Newton Method} \label{subsec:Newton}
We first present the Projected Quasi-Newton method which is proposed by \cite{schmidt2009optimizing}. We augment it with the algorithm proposed by \cite{wang2013projection} to project parameters onto a probability simplex in strongly polynomial time. In general, we address the problem of minimizing a differentiable function $ f(x)$ over a convex set $ \mathcal{C} $ subject to $ m $ equality constraints:
\begin{equation} \label{eqn:optimizeF}
\begin{array}{rl}
\min_{x} & f(x) \\
\mbox{s.t.} & h_j(x) \, = \, 0, \,\, \forall \, j=1,\cdots,m, \\
& x \, \in \, \mathcal C.
\end{array}
\end{equation}
In an iterative algorithm, we update the next iteration as follows:
\begin{equation}
x^{(t+1)} = x^{(t)} + \alpha_t d^{(t)},
\end{equation}
where $ x^{(t)} $ is the solution at the $t$-th iteration, $ \alpha_t $ is the step length and $ d^{(t)} $ is the moving direction at iteration $ t$. Different algorithms differ in how $d^{(t)}$ and $\alpha_t$ are determined. In the Projected Quasi-Newton method, a quadratic approximation of the objective function around the current iterate $ x^{(t)} $ is constructed as follows.
\begin{equation}\nonumber
q_t(x) = f(x^{(t)}) + (x - x^{(t)})^{T}g^{(t)} + \frac{1}{2}(x - x^{(t)})^{T}B^{(t)}(x - x^{(t)}),
\end{equation}
where $ g^{(t)} = \nabla f(x^{(t)}) $ and $ B^{(t)} $ denotes a positive-definite approximation of the Hessian $ \nabla^2 f(x^{(t)}) $. The projected quasi-Newton method then compute a feasible descent direction by minimizing this quadratic approximation subject to the original constraints:
\begin{equation} \label{eqn:optimizeQ}
\begin{array}{rl}
z^{(t)} \, = \, \mbox{argmin}_{x} & q_t(x),\\
\mbox{s.t.} &h_{j}(x) \, = \, 0, \,\,\forall\, j = 1, \cdots, m,\\
&x \, \in \, \mathcal{C}.
\end{array}
\end{equation}
Then the moving direction is $ d^{(t)} = z^{(t)}  - x^{(t)} $. \\

To determine the step length $ \alpha_t $, we ensure that a sufficient decrease condition, such as the Armijo condition is met:
\begin{equation} \label{eqn:Armijo}
f(x^{(t)}  + \alpha d_t) \, \le \, f(x^{(t)}) + \nu \alpha (g^{(t)})^{T}d^{(t)},
\end{equation}
where $ \nu \in (0, 1) $. \\

Although there are many appealing theoretical properties of projected Newton method just summarized, many obstacles prevent its efficient implementation in its original form. A major shortcoming is that minimizing (\oldref{eqn:optimizeQ}) could be as difficult as optimizing (\oldref{eqn:optimizeF}). In \cite{schmidt2009optimizing}, the projected Newton method was modified into a more practical version which uses the limited memory BFGS update to obtain $B^{(t)}$'s and a Spectral Projected Gradient (SPG) Algorithm (\citep{birgin2000nonmonotone}) to solve the quadratic approximation (\oldref{eqn:optimizeQ}).

To apply this Projected Quasi-Newton method to (\oldref{eqn:LMEasOptimization}), we let $f(\boldsymbol{\theta}) := -L(\boldsymbol\theta|Y)$. As we discussed in the previous section, we rewrite (\oldref{eqn:LMEasOptimization}) as follows:
\begin{equation}\label{eqn:LMEasOptimization_rewrite}
\begin{array}{ll}
\min & f(\boldsymbol{\theta}) \\
\mbox{s.t} & \boldsymbol{\theta} \in \mathcal F,
\end{array}
\end{equation}
where $\mathcal F = \mathcal P^K \otimes \bigotimes_{k=1}^K\bigotimes_{j=1}^d \mathcal P^{c_j}$ is the feasible region given in the format of the Cartesian product of $K\times d+1$ probability simplexes. This rewriting is to facilitate the projection operation. We denote $\Pi_{S}(x)$ as the projection of a vector $x\in \mathbb R^n$ on a closed convex set $S\subseteq \mathbb R^n$, i.e. $\Pi_{s}(x)$ is the unique solution of the following quadratic program:
\begin{equation}
\begin{array}{rl}
\min & \|y-x\|_2^2 \\
\mbox{s.t.} & y \in S.
\end{array}
\end{equation}
As we can see, in general, a quadratic program needs to be solved to compute the projection onto a closed convex set, and hence is not computationally cheap. Fortunately, the feasible region in (\oldref{eqn:LMEasOptimization_rewrite}) allows for a projection computable in strongly polynomial time according to \cite{wang2013projection}. This algorithm is presented in Algorithm \oldref{alg:projectionSimplex}. This algorithm is the building block for the SPG algorithm to solve the quadratic approximation in each iteration. More specifically, in the $t$-th iteration, let
\begin{equation}\nonumber
q_t(\boldsymbol{\theta}) = f(\boldsymbol{\theta}^{(t)}) + (\boldsymbol{\theta} - \boldsymbol{\theta}^{(t)})^{T}g^{(t)} + \frac{1}{2}(\boldsymbol{\theta} - \boldsymbol{\theta}^{(t)})^{T}B^{(t)}(\boldsymbol{\theta} - \boldsymbol{\theta}^{(t)}),
\end{equation}
where $ g^{(t)} = \nabla f(\boldsymbol{\theta}^{(t)}) $ and $ B^{(t)} $ denotes a positive-definite approximation of the Hessian $ \nabla^2 f(\boldsymbol{\theta}^{(t)})$. The quadratic approximation is now given by
\begin{equation} \label{eqn:optimizeQ2}
\begin{array}{rl}
\boldsymbol{\vartheta}{(t)} \, = \, \mbox{argmin}_{\boldsymbol{\theta}} & q_t(\boldsymbol{\theta}),\\
\mbox{s.t.} & \boldsymbol{\theta} \in \mathcal F.
\end{array}
\end{equation}
The gradient of $ q_t(\boldsymbol{\theta}) $ is given by
\begin{equation} \label{eqn:qk_derive}
\nabla q_t(\boldsymbol{\theta}) = \nabla f(\boldsymbol{\theta}^{(t)}) + (B^{(t)})^{T}(\boldsymbol{\theta} - \boldsymbol{\theta}^{(t)}).
\end{equation}
In our implementation, $ \nabla f(\boldsymbol{\theta}^{(t)}) $ is numerically approximated by the method of symmetric difference quotient with length chosen as $0.05$. We can also compute $\nabla f(\theta^{(t)})$ using the analytical expressions (\oldref{eqn:likelihood_gradient_eta}) and (\oldref{eqn:likelihood_gradient_pi}).

We update $ B^{(t)} $ using the limited memory version of BFGS. The non-limited memory BFGS update of $ B $ is given by
\begin{equation} \label{eqn:BFGS}
B^{(t+1)} = B^{(t)} - \frac{B^{(t)}s^{(t)}(s^{(t)})^{T}B^{(t)}}{(s^{(t)})^{T}B^{(t)}s^{(t)}} + \frac{y^{(t)}(y^{(t)})^{T}}{(y^{(t)})^{T}s^{(t)}},
\end{equation}
where $s^{(t)} = \boldsymbol{\theta}^{(t+1)} - \boldsymbol{\theta}^{(t)}$ and $y^{(t)} = \nabla f(\boldsymbol{\theta}^{(t+1)}) - \nabla f(\boldsymbol{\theta}^{(t)})$. This will consume significant memory in storing $ B^{(t)} $'s when the number of features increases dramatically. Therefore, in the proposed Projected Quasi-Newton algorithm we only keep the most recent $ m=5 $ $ Y $ and $ S $ arrays (the definitions of $ Y $ and $ S $ are in Algorithm \oldref{alg:projectedQN}) and update $B^{(t)}$ using its compact representation described by \cite{byrd1994representations}:
\begin{equation}
B^{(t)} \, = \, \sigma_t I - N^{(t)} (M^{(t)})^{-1} (N^{(t)})^{T},
\end{equation}
where $ N^{(t)} $ and $ M^{(t)} $ are explicitly given in equation (3.5) of \cite{byrd1994representations}. \\

In addition, running Algorithm \oldref{alg:projectedQN} until convergence, the $ B $ matrix is outputted as a by-product. The $ -B $ matrix is an approximation of the observed Fisher information of the unknown parameters, which will enable us to construct asymptotic confidence intervals using the following classical results:
\begin{equation}
\hat{\boldsymbol{\theta}} \rightarrow N(\boldsymbol{\theta}, -B_{\boldsymbol{\theta}}^{-1}).
\end{equation}

This is way easier than the EM algorithm to conduct statistical inference. According to \cite{gower2017randomized}, when $ f $ is convex quadratic function with positive definite Hessian matrix, it is expected that $ -B^{(t)} $ from the Quasi-Newton method to converge to the true Hessian matrix. However, the log-likelihood function is obviously not a convex function and as far as we know there is no formal theory that guarantees the convergence. Nonetheless, in Section~$ 6 $ of \cite{jamshidian1997acceleration}, the authors empirically compared the estimates for standard errors to the true values and the results are satisfactory.

In our implementation of Algorithm \oldref{alg:projectedQN}, we use $ m=5, \epsilon = 10^{-4} $ and the default parameters are $ \alpha_{\text{min}} = 10^{-10} $, $ \alpha_{\text{max}} = 10^{10} $, $ h = 1 $ and $ \nu = 10^{-4}$ in Algorithm \oldref{alg:SPG}. \\

\begin{algorithm}[htbp!]
	\caption{Limited Memory Projected Quasi-Newton Method}\label{alg:projectedQN}
	\begin{algorithmic}[1]
		\State Given $ \boldsymbol{\theta}^{(0)} $, $ m $ and $ \epsilon $. Set $ t = 0 $.
		\State \textbf{While} not converge:
		\State \indent $f^{(t)} = f(\boldsymbol{\theta}^{t}) $ and $ g^{(t)} = \nabla f(\boldsymbol{\theta}^{(t)}) $
		\State \indent Call \textbf{Algorithm} \oldref{alg:SPG} for $ \boldsymbol{\vartheta}^{(t)} $
		\State \indent $ d^{(t)} = \boldsymbol{\vartheta}^{(t)} - \boldsymbol{\theta}^{(t)} $
		\State \indent \textbf{If} $ \norm{ \Pi_{\mathcal F}(\boldsymbol{\theta}^{(t)}- g^{(t)}) - \boldsymbol{\theta}^{(t)}}_1 \le \epsilon $, where $ \Pi_{\mathcal F}(\cdot) $ calls \textbf{Algorithm} \oldref{alg:projectionSimplex}:
		\State \indent \indent Converged; \textbf{Break}.
		\State \indent $\alpha = 1 $
		\State \indent $ \boldsymbol{\theta}^{(t+1)} = \boldsymbol{\theta}^{(t)} + \alpha d^{(t)} $
		\State \indent \textbf{While} $ f(\boldsymbol{\theta}^{(t+1)}) > f^{(t)} + \nu \alpha (g^{(t)})^{T} d{(t)} $:
		\State \indent \indent Select $ \alpha $ randomly from Uniform distribution $ U(0, \alpha) $
        \State \indent \indent $ \boldsymbol{\theta}^{(t+1)} = \boldsymbol{\theta}^{(t)} + \alpha d^{(t)} $
		\State \indent $ s^{(t)} = \boldsymbol{\theta}^{(t+1)}  - \boldsymbol{\theta}^{(t)}  $
		\State \indent $ y^{(t)} = g^{(t+1)} - g^{(t)}  $
		\State \indent \textbf{If} $ t = 0 $:
		\State \indent \indent $ S = [s^{(t)}] $, $ Y = [y^{(t)}] $
		\State \indent \textbf{Else}:
		\State \indent \indent \textbf{If} $ t \ge m $:
		\State \indent \indent \indent Remove first column of $S$ and $Y$
		\State \indent \indent $S = [S, s^{(t)}] $
		\State \indent \indent $Y = [Y, y^{(t)}] $
		\State \indent $ \sigma^{(t)} = \frac{(y^{(t)})^{T}s^{(t)}}{(y^{(t)})^{T} y^{(t)}} $
		\State \indent Form $ N $ and $ M $ for BFGS update
		\State \indent $ t = t + 1  $
	\end{algorithmic}
\end{algorithm}

\begin{algorithm}[htbp!]
	\caption{Spectral Projected Gradient Algorithm}\label{alg:SPG}
	\begin{algorithmic}[1]
		\State Given $ x_0 $, step bounds $ 0 < \alpha_{\text{min}} < \alpha_{\text{max}} $
		\State Initial step length $\alpha_{bb} \in [\alpha_{\text{min}}, \alpha_{\text{max}}]$, and history length $ h $
		\State \textbf{While} not converge:
		\State \indent $\bar{\alpha}_k = \min\{ \alpha_{\text{max}}, \max\{ \alpha_{\text{min}}, \alpha_{bb}\} \} $
		\State \indent $d_k = \mathcal{P}_c(x_k - \bar{\alpha}_k \nabla q_k(x_k)) - x_k $, where $ \mathcal{P}_c(.) $ calls \textbf{Algorithm} \oldref{alg:projectionSimplex}.
		\State \indent Set bound $f_b = \max\{ f(x_k), \cdots, f(x_{k - h}) \}$
		\State \indent $ \alpha = 1  $
		\State \indent \textbf{While} $ q_k(x_k + \alpha d_k) > f_b + \nu \alpha \nabla q_k(x_k)^{T} d_k$:
		\State \indent \indent Select $ \alpha $ randomly from Uniform distribution $ U(0, \alpha) $.
		\State \indent $ x_{k+1} = x_k + \alpha d_k $
		\State \indent $ s_{k} = x_{k+1} - x_k $
		\State \indent $ y_k = \nabla q_k(x_{k+1}) - \nabla q_k(x_k) $
		\State \indent $ \alpha_{bb} = y_k^{T}y_k/s_k^{T}s_k $
		\State \indent $ k = k + 1  $	
	\end{algorithmic}
\end{algorithm}

\begin{algorithm}[htbp!]
	\caption{Euclidean Projection of a Vector onto the Probability Simplex.}\label{alg:projectionSimplex}
	\begin{algorithmic}[1]
		\State Supply $\boldsymbol{x} \in R^{D}$.
		\State Sort $ \boldsymbol{x} $ into $ \boldsymbol{u} $ such that $ u_1 \ge u_2 \ge \cdots \ge u_D $
		\State Find $ \rho = \max\{ 1 \le j \le D: u_j + \frac{1}{j}(1 - \sum_{i=1}^{j} u_i) > 0 \} $
		\State Define $ \lambda = \frac{1}{\rho} (1 - \sum_{i=1}^{\rho} u_i) $
		\State Output $ \boldsymbol{x}^{\prime} $ such that $  x_i^{\prime} = \max\{x_i + \lambda, 0\}, i = 1, \cdots, D$.
	\end{algorithmic}
\end{algorithm}

\subsection{Sequential Quadratic Programming} \label{subsec:SQP}

Sequential quadratic programming (SQP) is a generic method for non-linear optimization with constraints. It is known as one of the most efficient computational method to solve the general nonlinear programming problem in (\oldref{eqn:optimizeF}) subject to both equality and inequality constraints. There are many variants of this algorithm, we use the version considered in \cite{kraft1988software}. We give a brief review of this method and then we will specifically talk about how this method could be applied to optimization problem (\oldref{eqn:LMEasOptimization}). \\

Consider the following minimization problem
\begin{equation}
\begin{array}{rl}
\displaystyle{\min_{x} } & f(x)\\[5pt]
\mbox{s.t.} & c_{j}(x) \, = \, 0, j = 1, 2, \cdots, m_e,\\
& c_{j}(x) \, \geq\,  0, j = m_e+1, m_e+2, \cdots, m,\\
& x_l \, \le \, x \, \le \, x_u,
\end{array}
\end{equation}
where the problem functions $ f: \mathbb R^{n} \rightarrow \mathbb R$. SQP is also an iterative method and each iteration a quadratic approximation of the original problem is also constructed and solved to obtain the moving direction. Compared to the Projected Quasi-Newton method, in SQP, the quadratic approximations are typically solved by an active set method or an interior point method rather than a projection type method. This significantly complicates the algorithm, but also allows the algorithm to handle more general non-linear optimization problems, especially when the feasible region is too complex to admit an efficient projection computation. In particular, starting with a given vector of parameters $x^{(0)} $, the moving direction $ d^{(t)} $ at iteration $t$  is determined by a quadratic programming problem, which is formulated by a quadratic approximation of the Lagrangian function and a linear approximation of the constraints. Note that, in contrast to the Projected Quasi-Newton method we presented in the previous subsection, the SQP algorithm here approximates the Lagrangian function instead of the objective function itself. An advantage is that the dual information can be incorporated in the algorithm to ensure better convergence property. Let
\begin{equation} \label{eqn:Lagrange}
L(x; \lambda) \, = \, f(x) - \sum_{j=1}^{m} \lambda_j c_j(x),
\end{equation}
be the Lagrangian function associated with this optimization problem. This approximation is of the following standard form of quadratic programming:
\begin{equation} \label{eqn:SQP_QP}
\begin{array}{rl}
\displaystyle{\min_{x}} & \displaystyle{\frac{1}{2}} (x - x^{(t)})^{T}B^{(t)}(x - x^{(t)}) + \nabla f(x^{(t)})(x - x^{(t)})\\[5pt]
\mbox{s.t.} & (\nabla c_j(x^{(t)}))^T(x- x^{(t)}) + c_j(x^{(t)}) \, = \, 0, j = 1, 2, \cdots, m_e, \\
&(\nabla c_j(x^{(t)}))^T(x- x^{(t)}) + c_j(x^{(t)}) \, \ge\, 0, j = m_e+1, m_e+2, \cdots, m,\\
\end{array}
\end{equation}
with
\begin{equation} \label{eqn:B}
B^{(t)} = \nabla_{xx}^2 L(x^{(t)},\lambda^{(t)}),
\end{equation}
as proposed in \cite{wilson1963simplicial}. The multiplier $\lambda^{(t)}$ is updated using the multipliers of the constraints in (\ref{eqn:SQP_QP}).

In terms of the step length $ \alpha $, \cite{han1977globally} proved that a one-dimensional minimization of the non-differential exact penalty function
\begin{equation*}
\phi(x; \varrho) = f(x) + \sum_{j=1}^{m_e} \varrho_j|c_j(x)| + \sum_{j=m_e + 1}^{m} \varrho_j|c_j(x)|_{-}
\end{equation*}
with $ |c_j(x)|_{-} = | \min \left(0; c_j(x)\right)| $, as a merit function $ \varphi: \mathbb R \rightarrow \mathbb R $
\begin{equation*}
\varphi(\alpha) = \phi(x^{(t)} + \alpha d^{(t)}),
\end{equation*}
with $ x^{(t)} $ and $ d^{(t)} $ fixed, leads to a step length $ \alpha $ guaranteeing global convergence for values of the penalty parameters $ \varrho_j $ greater than some lower bounds. Then, \cite{powell1978fast} proposed to update the penalty parameters according to
\begin{equation*}
\varrho_j = \max( \frac{1}{2}(\varrho_j^{-} + |\mu_j|), |\mu_j|), j = 1, \cdots, m,
\end{equation*}
where $ \mu_j $ denotes the Lagrange multiplier of the $j$-th constraint in the quadratic problem and $ \varrho_j^{-}$ is the $j$-th penalty parameter of the previous iteration, starting with some $ \varrho_j^{0} =0$.\\

It is important in practical applications to not evaluate $ B^{(t)} $ in (\oldref{eqn:B}) in every iteration, but to use only first order information to approximate the Hessian matrix of the Lagrange function in (\oldref{eqn:Lagrange}). \cite{powell1978fast} proposed the following modification:
\begin{equation*}
B^{(t+1)} = B^{(t)} + \frac{q^{(t)} (q^{(t)})^{T}}{(q^{(t)})^T s^{(t)}} - \frac{B^{(t)} s^{(t)} (s^{(t)})^{T}B^{(t)}}{(s^{(t)})^{T}B^{(t)} s^{(t)}},
\end{equation*}
with
\begin{eqnarray}
s^{(t)} &=& x^{(t+1)} - x^{(t)} \nonumber \\
q^{(t)} &=& \gamma_t \eta^{(t)} + (1 - \gamma_t) B^{(t)} s^{(t)} \nonumber
\end{eqnarray}
where
\begin{equation*}
\eta^{(t)} = \nabla_{x} L(x^{(t+1)}, \lambda^{(t)}) - \nabla_{x} L(x^{(t)}, \lambda^{(t)})
\end{equation*}
and $ \gamma_t $ is chosen as
\begin{equation*}
\gamma_t =
\begin{cases}
1 & \text{if}~ (s^{(t)})^{T} \eta^{(t)} \ge 0.2 (s^{(t)})^{T}B^{(t)} s^{(t)}, \\
\frac{0.8 (s^{(t)})^{T} B^{(t)} s^{(t)}}{(s^{(t)})^{T}B^{(t)} s^{(t)} - (s^{(t)})^{T}\eta^{(t)}} & \text{otherwise},
\end{cases}
\end{equation*}
which ensures that $ B^{(t+1)} $ remains positive definite within the linear manifold defined by the tangent planes to active constraints at $ x^{(t+1)} $. \\

In LCM, the problem turns out to be simpler: the quadratic programming problem in (\oldref{eqn:SQP_QP}) is only subject to $ m_e $ equality constraints. In addition, unless we use Projected Quasi-Newton method, for which we have to build our own solver, there is a popular implementation of SQP in Python's SciPy package. The package uses a variant of SQP: Sequential Least SQuares Programming (SLSQP): It replaces the quadratic programming problem in (\oldref{eqn:SQP_QP}) by a linear least squares problem using a stable $ LDL^{T}$ factorization of the matrix $B^{(t)}$.

\section{Simulation Studies and Real Data Analysis} \label{sec:sim}
In this section, we provide four example bundles and one real data analysis to demonstrate the performance of the proposed methods. The model specifications of the four example bundles as follows:
	\begin{itemize}
		\item Example Bundle~$1$, $ N=500 $: (A) $ d=1, K=2 $; (B) $ d=1, K=3 $; (C) $ d=2, K=2 $; (D) $ d=4, K=2 $
		\item Example Bundle~$2$, $ N=1000 $: (A) $ d=2, K=2 $; (B) $ d=2, K=3 $; (C) $ d=3, K=2 $; (D) $ d=3, K=3 $
		\item Example Bundle~$3$, $ N=2000 $: (A) $ d=3, K=3 $; (B) $ d=3, K=4 $; (C) $ d=4, K=4 $; (D) $ d=5, K=3 $
		\item Example Bundle~$4$, $ N=5000 $: (A) $ d=4, K=4 $; (B) $ d=4, K=5 $; (C) $ d=5, K=4 $; (D) $ d=5, K=5 $
	\end{itemize}
One dataset is simulated from latent class model for each combination. In total, we consider $ 16 $ datasets with different combinations of sample size, dimensionality and number of groups providing a comprehensive picture of the model performance.

\subsection{Example Bundle~1} \label{subsec:example1}
\begin{figure}[htbp!]
	\centering
	\hspace*{-0.5cm}
	\includegraphics[width=1.1\textwidth, angle = 0]{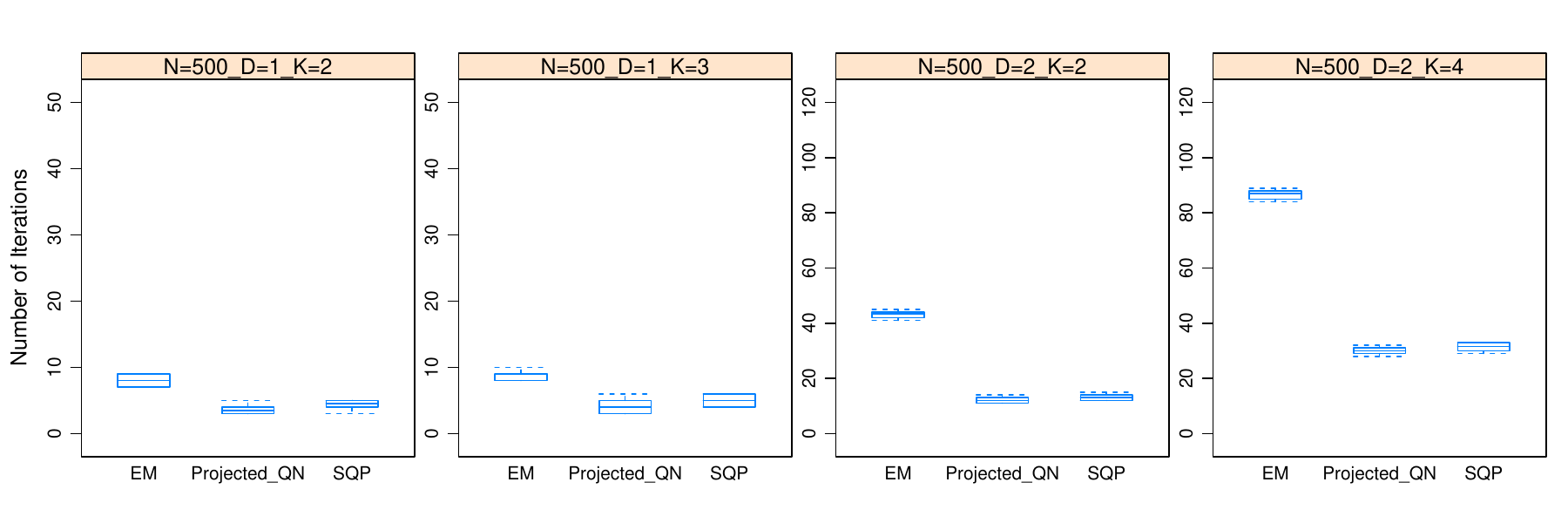}
	\caption{ Example Bundle $1$, $ N=500 $; number of iterations for (A) $ d=1, K=2 $; (B) $ d=1, K=3 $; (C) $ d=2, K=2 $; (D) $ d=4, K=2 $.}
	\label{plot:b1:ir}
\end{figure}

\begin{figure}[htbp!]
	\centering
	\hspace*{-0.5cm}
	\includegraphics[width=1.1\textwidth, angle = 0]{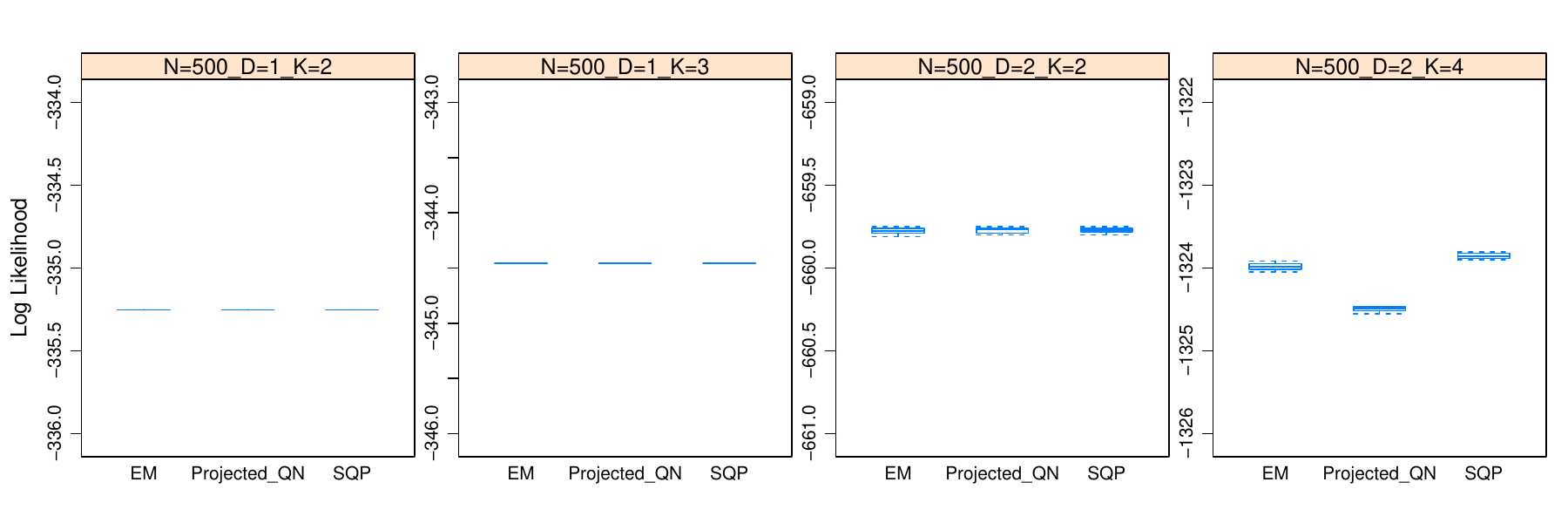}
	\caption{Example Bundle $1$, $ N=500 $; log-likelihood values for (A) $ d=1, K=2 $; (B) $ d=1, K=3 $; (C) $ d=2, K=2 $; (D) $ d=4, K=2 $.}
	\label{plot:b1:lk}
\end{figure}

In this example bundle, we use the following three methods to maximize the log-likelihood function: (1) EM, (2) SQP, and (3) Projected Quasi-Newton (QN). Each method is repeated $ 10 $ times with different initial values across the $ 10 $ runs. At each run, the three methods begin with identical initial values. The true weights and categorical parameters are reported in Tables~\oldref{table:b1:1}, \oldref{table:b1:2}, \oldref{table:b1:3}, \oldref{table:b1:4} in the appendix. Side by side boxplots are drawn and reported in Figure~\oldref{plot:b1:ir} and Figure~\oldref{plot:b1:lk} showing number of iterations and log-likelihood values of the $ 10 $ runs, respectively. For each method, the best result based on the log-likelihood values across the $ 10 $ runs are given in Table~\oldref{table:b1}. Results from the true parameters are also included in Table~\oldref{table:b1} as a comparison.

\begin{table}[htbp!]
	\caption{ Example Bundle $1$, $ N=500 $; the best result based on the log-likelihood among the $ 10 $ runs for each method.}
	\hspace*{-0.5cm}
		\begin{tabular}{|c|c | c |  c |  c|}
			 \multicolumn{5}{c}{(A) $ d=1, K=2 $} \\ \hline
		& True Parameters &  EM & SQP & Projected QN      \\ \hline
			Log-Likelihood & $-335.29$ & $-335.25$ & $-335.25$ & $-335.25$     \\ \hline
			Number of Iterations & N.A. & $8$ & $4$ & $4$  \\ \hline
			
		 \multicolumn{5}{c}{(B) $ d=1, K=3 $} \\ \hline		
& True Parameters &  EM & SQP & Projected QN      \\ \hline
	 Log-Likelihood & $-344.49$ & $-344.45$  & $-344.45$ & $-344.45$    \\ \hline
	 Number of Iterations & N.A. &  $9$  & $5$  & $4$  \\ \hline		
	
	 \multicolumn{5}{c}{(C) $ d=2, K=2 $} \\ \hline
	& True Parameters &  EM & SQP & Projected QN  \\ \hline
	Log-Likelihood & $-661.30$ & $-659.75$ & $-659.75$ & $-659.75$   \\ \hline
	Number of Iterations & N.A. & $43$ & $13$ & $12$  \\ \hline			
	
	 \multicolumn{5}{c}{(D) $ d=4, K=2 $} \\ \hline
	& True Parameters &  EM & SQP & Projected QN      \\ \hline
	Log-Likelihood  & $ -1323.29 $   & $-1323.91$  & $-1323.80$ & $-1324.46$    \\ \hline
	Number of Iterations  & N.A. & $88$  & $31$  & $30$  \\ \hline	
				
		\end{tabular}
	\label{table:b1}
\end{table}

From Table~\oldref{table:b1}, Figure~\oldref{plot:b1:ir} and Figure~\oldref{plot:b1:lk}, we observe that the proposed two optimization methods have good performance compared to the traditional EM algorithm: the log-likelihood values are very close to that of EM for all four datasets in this example bundle. Note that the vertical axis scales are different in Figure~\oldref{plot:b1:ir}. The numbers of iterations of the two proposed optimization methods are obviously lower than that of EM, for example the number of iterations of SQP and Projected QN are both $ 12 $ compared to $ 88 $ of the EM algorithm. This suggests that the two optimization methods are less likely to get stuck in local maxima. \\

In addition, there are no substantial differences between the final best solutions across the $10$ runs. Actually, the final best results are quite close to the results obtained from the other $9$ runs. Using scenario (A) with $ d=1, K=2 $ in this bundle as an example, we divide the $ 10 $ log-likelihood values into two groups, where the first group contains the largest log-likelihood value only while the second group contains the rest of the nine log-likelihood values, and then fit a non-parametric two-group Wilcoxon signed-rank test \cite{wilcox}. The p-value is $ 0.20 $, which is clearly larger than the usual $ 0.05 $ threshold. The parametric t-test might not work well here because the group sizes are too small. Moreover, the estimated weights and categorical parameters from the $10$ runs are also close to each other. We repeat the test for the log-likelihood values on the estimates for each of the weight and categorical parameters and none of the p-values are larger than $ 0.05$.

\subsection{Example Bundle~2}
\begin{figure}[htbp!]
	\centering
	\hspace*{-0.5cm}
	\includegraphics[width=1.1\textwidth, angle = 0]{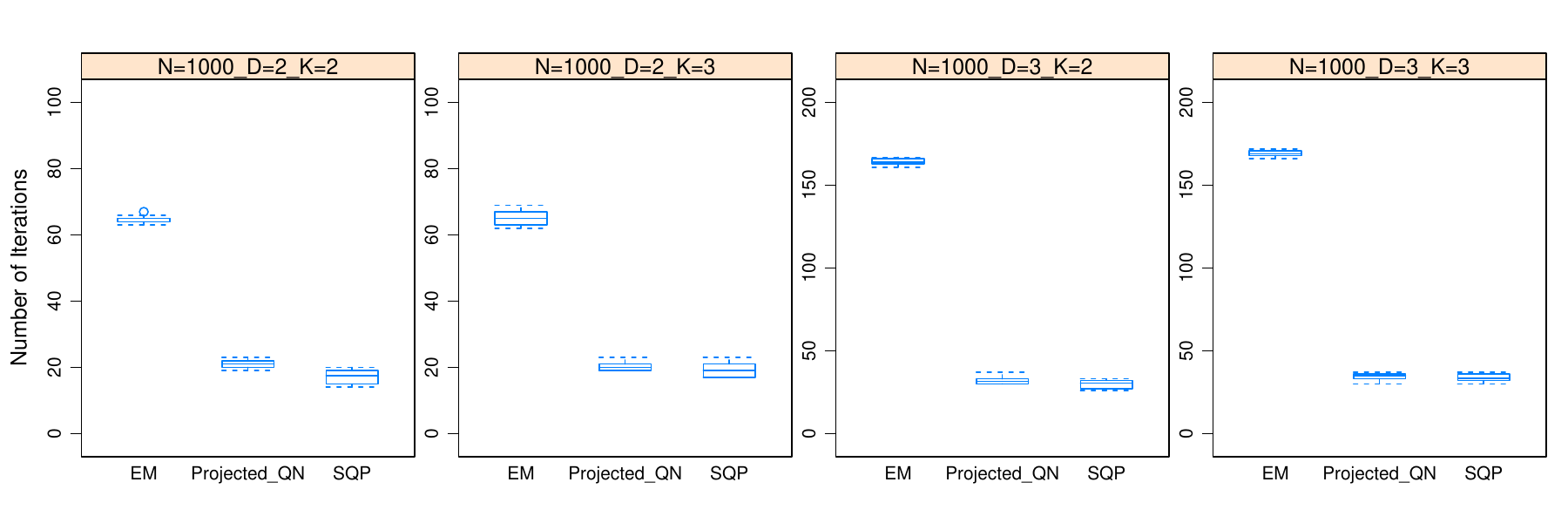}
	\caption{ Example Bundle $2$, $N=1000$; number of iterations for  (A) $ d=2, K=2 $; (B) $ d=2, K=3 $; (C) $ d=3, K=2 $; (D) $ d=3, K=3 $.}
	\label{plot:b2:ir}
\end{figure}

\begin{figure}[htbp!]
	\centering
	\hspace*{-0.5cm}
	\includegraphics[width=1.1\textwidth, angle = 0]{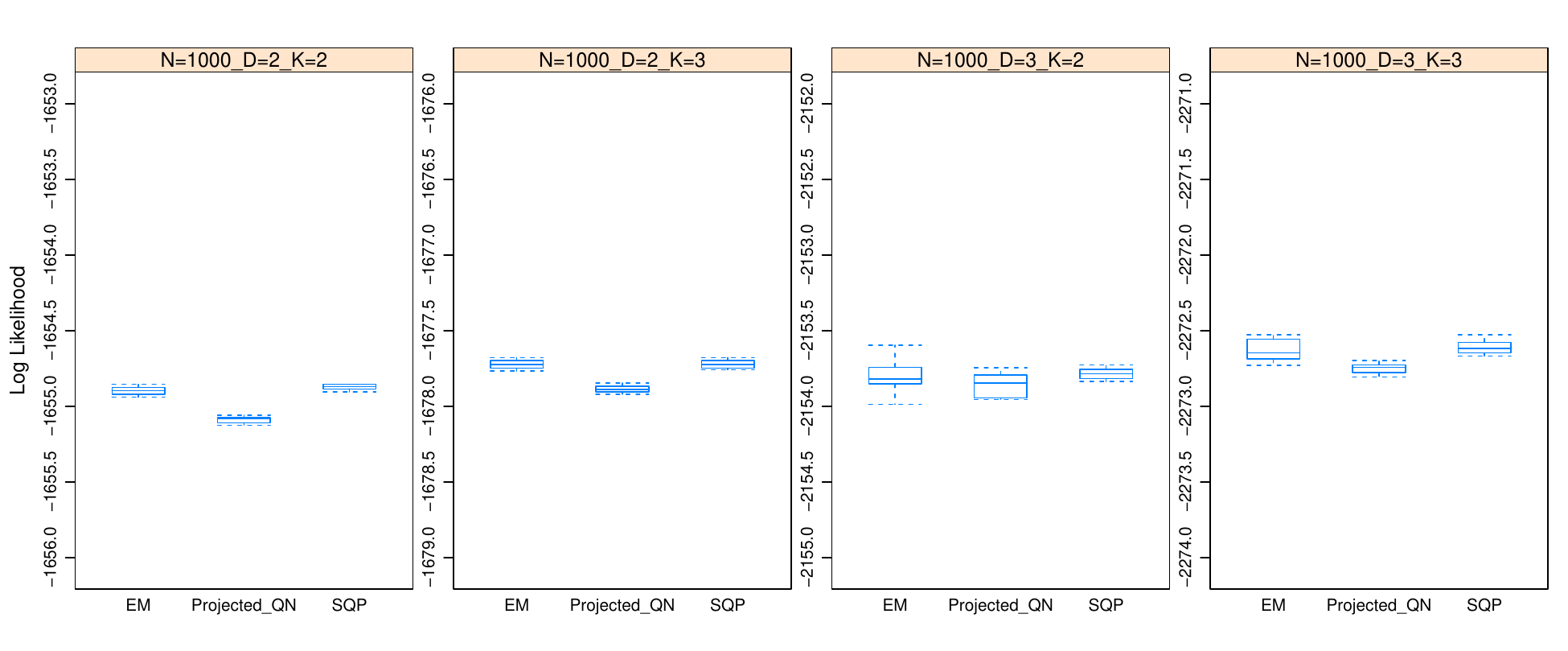}
	\caption{Example Bundle $2$, $N=1000$; log-likelihood values for  (A) $ d=2, K=2 $; (B) $ d=2, K=3 $; (C) $ d=3, K=2 $; (D) $ d=3, K=3 $.}
	\label{plot:b2:lk}
\end{figure}

The true weights and categorical parameters are reported in Tables~\oldref{table:b2:1}, \oldref{table:b2:2}, \oldref{table:b2:3}, \oldref{table:b2:4} in the appendix. As in Example Bundle~1, each method is repeated $ 10 $ times with different initial values across the $ 10 $ runs. At each run, the three methods begin with identical initial values. The simulation results for this bundle are summarized in Figure~\oldref{plot:b2:ir} and \oldref{plot:b2:lk} for number of iterations and log-likelihood, respectively. Similarly, for each method, the best result based on the log-likelihood values among the $ 10 $ runs are given in Table~\oldref{table:b2}. Results from the true parameters are also included in Table~\oldref{table:b2} for comparison.

\begin{table}[htbp!]
	\caption{Example Bundle $2$, $N=1000$; the best result based on the log-likelihood values among the $ 10 $ runs for each method.}
	\hspace*{-0.5cm}
		\begin{tabular}{|l|c c  c  c|}
			 \multicolumn{5}{c}{(A) $ d=2, K=2 $}  \\ \hline
			& True Parameters &  EM & SQP & Projected QN      \\ \hline
			Log-Likelihood & $-1656.59$ & $-1654.86$ & $-1654.86$ & $-1655.06$     \\ \hline
			Number of Iterations & N.A. & $65$ & $17$ & $21$   \\ \hline		
			
		  \multicolumn{5}{c}{(B) $ d=2, K=3 $} \\ \hline
		& True Parameters &  EM & SQP & Projected QN       \\ \hline
		Log-Likelihood  & $-1679.85$ & $-1677.68$  & $-1677.68$ & $-1677.85$    \\ \hline
		Number of Iterations  & N.A. &  $75$  & $21$  & $22$  \\ \hline		

			 \multicolumn{5}{c}{(C) $ d=3, K=2 $}  \\ \hline
			& True Parameters &  EM & SQP & Projected QN       \\ \hline
			Log-Likelihood & $-2156.11$ & $-2153.60$ & $-2153.73$ & $-2153.74$     \\ \hline
			Number of Iterations & N.A. & $165$ & $29$ & $32$   \\ \hline

		 \multicolumn{5}{c}{(D) $ d=3, K=3 $} \\ \hline
		& True Parameters &  EM & SQP & Projected QN      \\ \hline
		Log-Likelihood  & $ -2274.72 $   & $-2272.53$  & $-2272.53$ & $-2272.70$    \\ \hline
		Number of Iterations & N.A. & $169$  & $34$  & $35$  \\ \hline			
	\end{tabular}
	\label{table:b2}
\end{table}

From Table~\oldref{table:b2}, Figure~\oldref{plot:b2:ir} and Figure~\oldref{plot:b2:lk}, we observed a similar pattern as in Example Bundle~1, i.e., the log-likelihood values are close to each other, however the number of iterations of the two optimization methods are smaller than that of EM, further showing the promise of using the proposed optimization methods as alternatives in practice. \\

\subsection{Example Bundle~3} \label{subsec:example3}
\begin{figure}[htbp!]
	\centering
	\hspace*{-0.5cm}
	\includegraphics[width=1.1\textwidth, angle = 0]{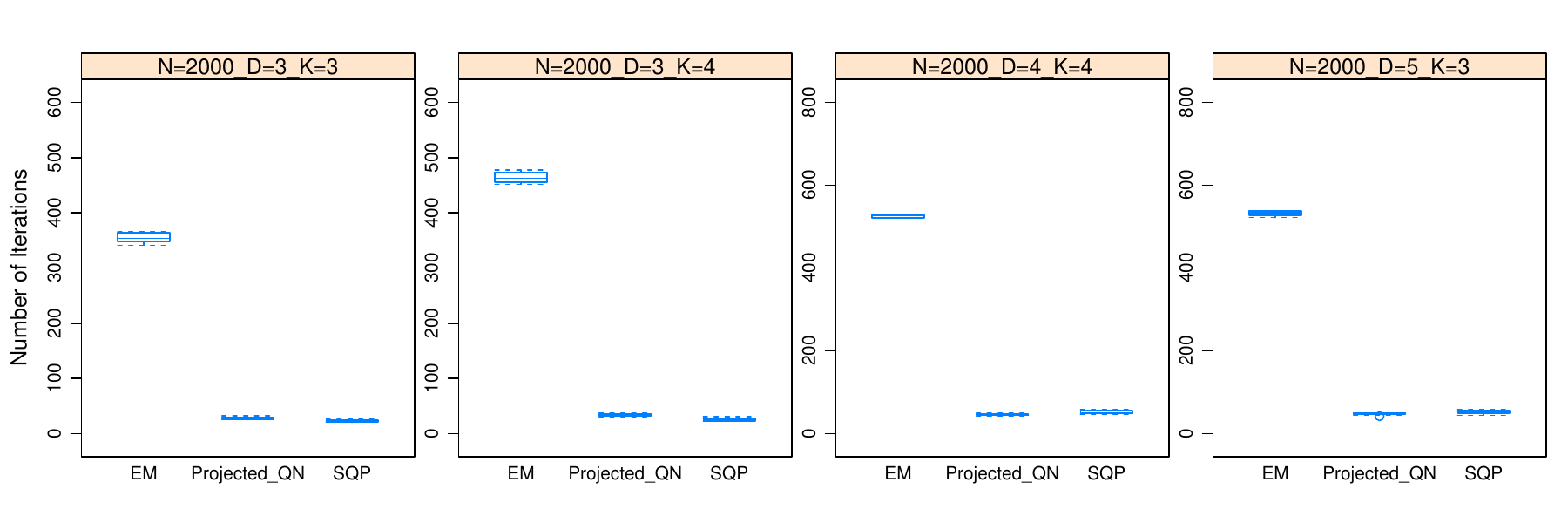}
	\caption{Example Bundle $3$, $N=2000$; number of iterations for (A) $ d=3, K=3 $; (B) $ d=3, K=4 $; (C) $ d=4, K=4 $; (D) $ d=5, K=3 $.}
	\label{plot:b3:ir}
\end{figure}

\begin{figure}[htbp!]
	\centering
	\hspace*{-0.5cm}
	\includegraphics[width=1.1\textwidth, angle = 0]{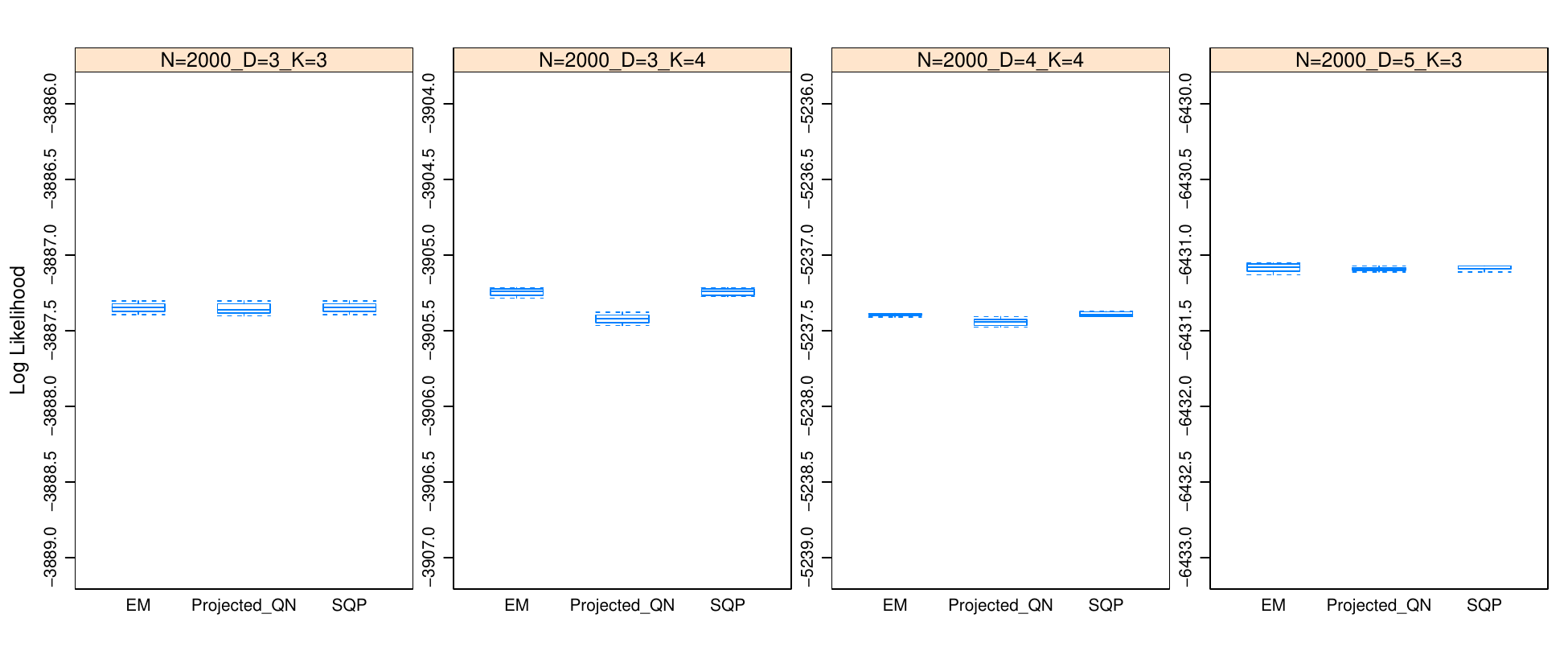}
	\caption{Example Bundle $3$, $N=2000$; log-likelihood values for (A) $ d=3, K=3 $; (B) $ d=3, K=4 $; (C) $ d=4, K=4 $; (D) $ d=5, K=3 $.}
	\label{plot:b3:lk}
\end{figure}

With exactly the same settings, we report results for Example Bundle~3 in this section. The resulting number of iteration and log-likelihood values are reported in Figure~\oldref{plot:b3:ir} and \oldref{plot:b3:lk}, respectively. For each method, the best result based on the log-likelihood values among the $ 10 $ runs are given in Table~\oldref{table:b3}. Results from the true parameters are also included in Table~\oldref{table:b3} for comparison. The true weights and categorical parameters are reported in  Tables~\oldref{table:b3:1}, \oldref{table:b3:2}, \oldref{table:b3:3}, \oldref{table:b3:4} in the appendix.

\begin{table}[htbp!]
	\caption{Example Bundle $3$, $N=2000$; the best result based on the log-likelihood values among the $ 10 $ runs for each method.}
	\hspace*{-0.5cm}
		\begin{tabular}{|l|c c  c  c|}
			 \multicolumn{5}{c}{(A) $ d=3, K=3 $}  \\ \hline
			& True Parameters &  EM & SQP & Projected QN      \\ \hline
			Log-Likelihood & $-3888.77$ & $-3887.30$ & $-3887.30$ & $-3887.30$     \\ \hline
			Number of Iterations & N.A. & $355$ & $23$ & $28$   \\ \hline		

		 \multicolumn{5}{c}{(B) $ d=3, K=4 $}  \\ \hline
		& True Parameters &  EM & SQP & Projected QN       \\ \hline
		Log-Likelihood  & $-3906.01$ & $-3905.22$  & $-3905.22$ & $-3905.38$    \\ \hline
		Number of Iterations  & N.A. &  $464$  & $26$  & $34$  \\ \hline		

			 \multicolumn{5}{c}{(C) $ d=4, K=3 $}  \\ \hline
			& True Parameters &  EM & SQP & Projected QN       \\ \hline
			Log-Likelihood & $-5241.99$ & $-5237.39$ & $-5237.37$ & $-5237.41$     \\ \hline
			Number of Iterations & N.A. & $526$ & $51$ & $46$   \\ \hline			

		  \multicolumn{5}{c}{(D) $ d=5, K=3 $} \\ \hline
		 & True Parameters &  EM & SQP & Projected QN      \\ \hline
		Log-Likelihood  & $ -6437.05 $   & $-6431.05$  & $-6431.07$ & $-6431.07$    \\ \hline
		Number of Iterations  & N.A. & $533$  & $53$  & $48$  \\ \hline			
	\end{tabular}
	\label{table:b3}
\end{table}

From Table~\oldref{table:b3}, Figure~\oldref{plot:b3:ir} and Figure~\oldref{plot:b3:lk}, we observed a similar pattern as in the previous examples: the number of iterations of the two optimization methods are much smaller than that of EM, while the log-likelihood values are quite close to each other for the three methods.

\subsection{Example Bundle 4}
\begin{figure}[htbp!]
	\centering
	\hspace*{-0.5cm}
	\includegraphics[width=1.1\textwidth, angle = 0]{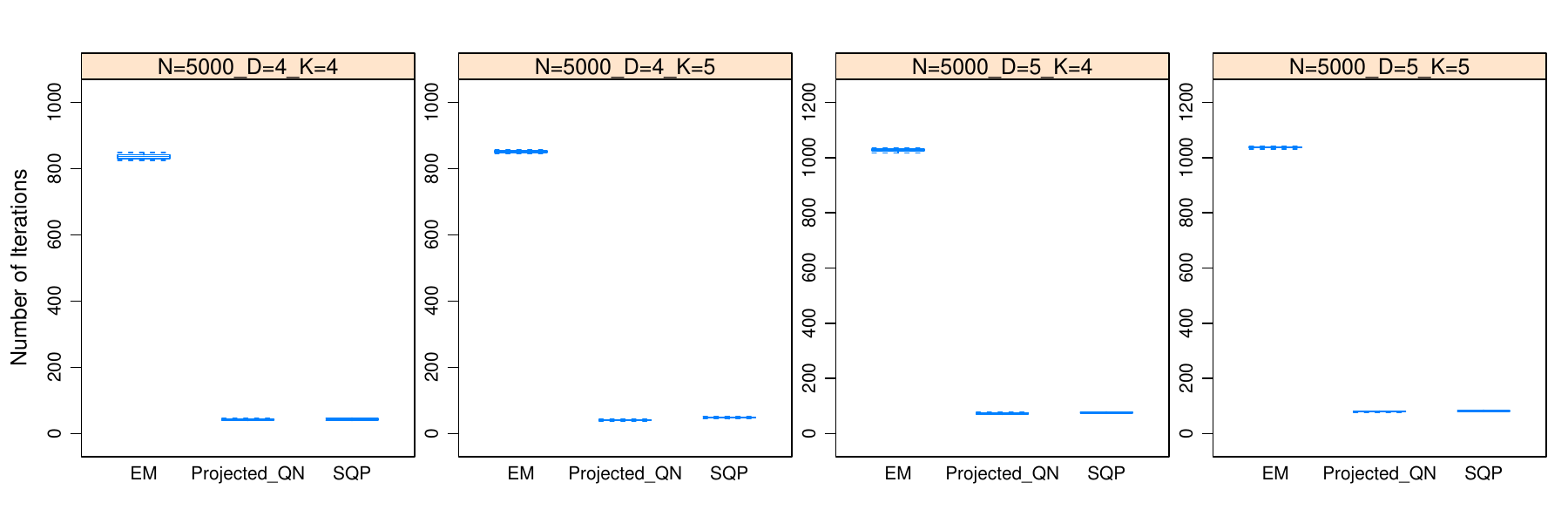}
	\caption{Example Bundle $4$, number of iterations.}
	\label{plot:b4:ir}
\end{figure}

\begin{figure}[htbp!]
	\centering
	\hspace*{-0.5cm}
	\includegraphics[width=1.1\textwidth, angle = 0]{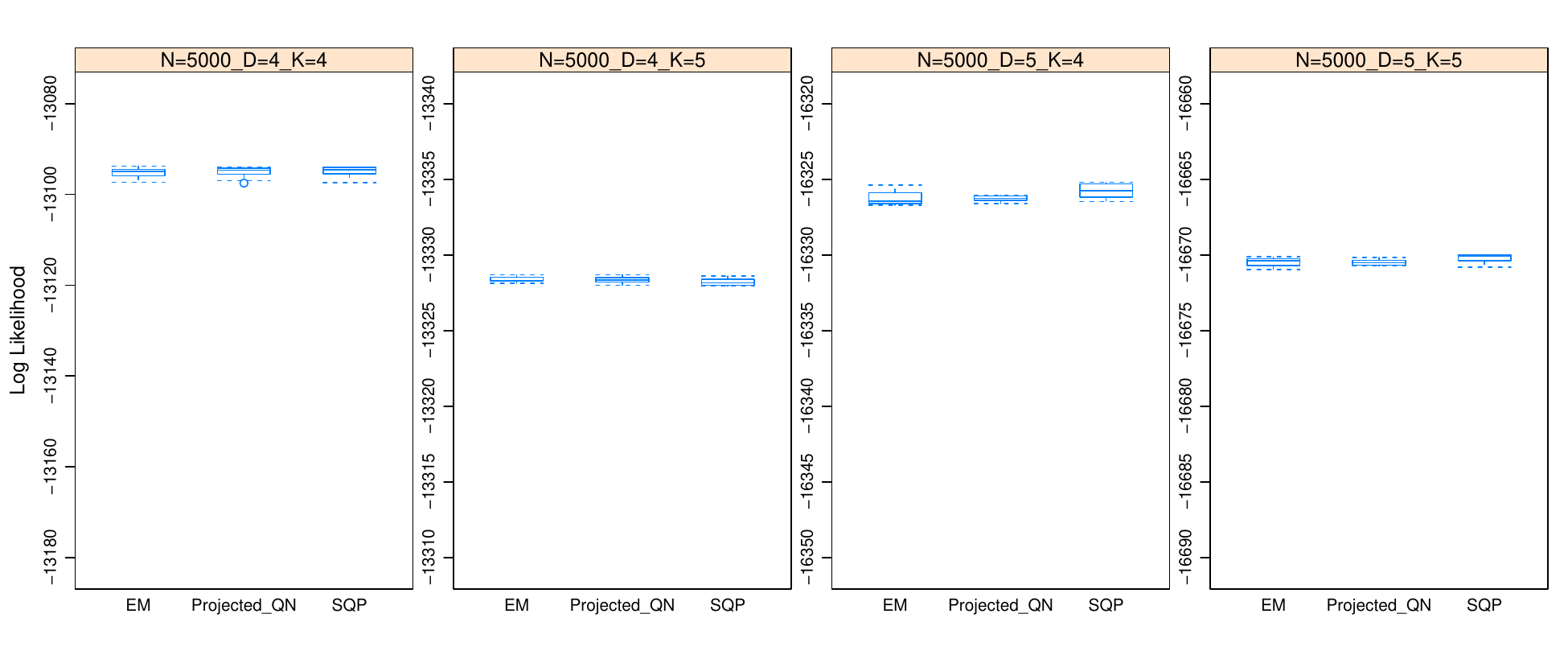}
	\caption{Example Bundle $4$, log-likelihood.}
	\label{plot:b4:lk}
\end{figure}

The resulting number of iterations and log-likelihood values of Example Bundle~$4$ are reported in Figure~\oldref{plot:b4:ir} and \oldref{plot:b4:lk}, respectively. For each method, the best result based on the log-likelihood values among the ten runs are given in Table~\oldref{table:b4}. Results from the true parameters are also included in Table~\oldref{table:b4} for comparison. The true weights and categorical parameters are repeated in Tables~\oldref{table:b4:1}, \oldref{table:b4:2}, \oldref{table:b4:3}, \oldref{table:b4:4} in the appendix.

\begin{table}[htbp!]
	\caption{Example Bundle $4$, the best result based on the log-likelihood values among the $ 10 $ runs for each method.}
	\hspace*{-0.5cm}
	\begin{tabular}{|l|c c  c  c|}
		\multicolumn{5}{c}{(A) $ d=4, K=4 $}  \\ \hline
		& True Parameters &  EM & SQP & Projected QN      \\ \hline
		Log-Likelihood & $-13105.82$ & $-13093.78$ & $-13093.98$ & $-13093.92$     \\ \hline
		Number of Iterations & N.A. & $837$ & $43$ & $42$   \\ \hline		
		
		\multicolumn{5}{c}{(B) $ d=4, K=5 $}  \\ \hline
		& True Parameters &  EM & SQP & Projected QN       \\ \hline
		Log-Likelihood  & $-13335.47$ & $-13328.15$  & $-13327.94$ & $-13328.02$    \\ \hline
		Number of Iterations  & N.A. &  $852$  & $48$  & $40$  \\ \hline		
		
		\multicolumn{5}{c}{(C) $ d=5, K=4 $}  \\ \hline
		& True Parameters &  EM & SQP & Projected QN       \\ \hline
		Log-Likelihood & $-16336.15$  & $-16325.38$ & $-16325.18$ & $-16326.06$     \\ \hline
		Number of Iterations & N.A. & $1028$ & $76$ & $73$   \\ \hline			
		
		\multicolumn{5}{c}{(D) $ d=5, K=5 $} \\ \hline
		& True Parameters &  EM & SQP & Projected QN      \\ \hline
		Log-Likelihood  & $-16684.59$   & $-16670.12$  & $-16669.97$ & $-16670.17$    \\ \hline
		Number of Iterations  & N.A. & $1038$  & $82$  & $80$  \\ \hline			
	\end{tabular}
	\label{table:b4}
\end{table}

These results in Example Bundle~4 further confirm what we have observed: with the same settings, the two optimization methods converge in less iterations than EM, while they still yield comparable log-likelihood values as EM. This strengthens the promise of using the two proposed optimization methods as alternatives to EM when estimating a latent class model.

\subsection{An Application}
\begin{figure}[htbp!]
	\centering
	\begin{subfigure}{.5\textwidth}
		\centering
		\includegraphics[width=0.9\linewidth]{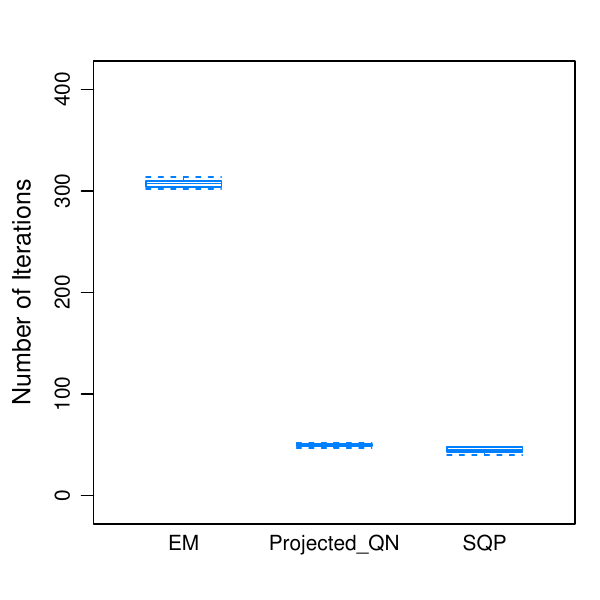}
		\caption{Number of Iterations}
		\label{fig:sub1}
	\end{subfigure}%
	\begin{subfigure}{.5\textwidth}
		\centering
		\includegraphics[width=0.9\linewidth]{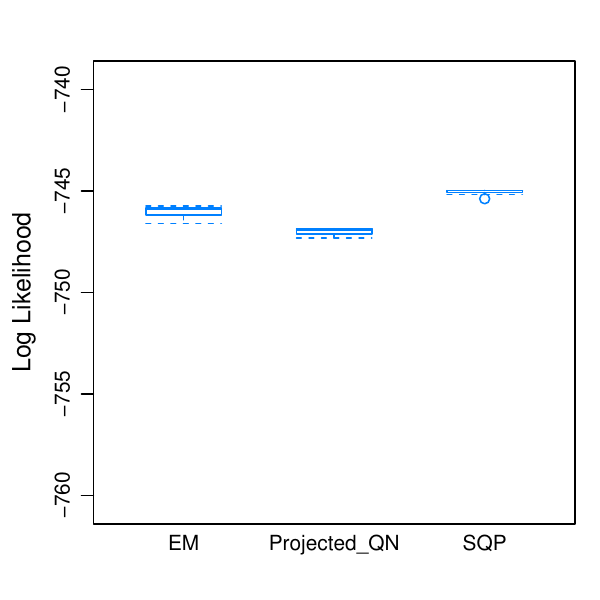}
		\caption{Log-Likelihood}
		\label{fig:sub2}
	\end{subfigure}
	\caption{Boxplots for the application example.}
	\label{plot:app:ir}
\end{figure}

We now go back to the motivating example discussed in Section~\oldref{sec:intro}. The data set is available in the R package \textit{BayesLCA}. \cite{white2014bayeslca} used a $ K =3 $ latent class model to fit the data using Gibbs sampler. It is clear that this is an $ n = 240, d = 6 $ binary data set. We follow the recommendation of \cite{moran2004syndromes} and fit a $ K = 3 $ latent class model with (1) EM, (2) SQP, and (3) Projected Quasi-Newton methods and $ 10$ different initial points, and the best result of each method is recorded based on the log-likelihood value. The results are summarized in Table~\oldref{table:app_results}. The result from BayesLCA package \citep{white2014bayeslca} is also included. The side-by-side boxplots for number of iterations and log-likelihood values of the $10$ runs are reported in Figure~\oldref{plot:app:ir}.

\begin{table}[htbp!]
	\captionsetup{labelfont={color=black}}
	\caption{Performance of the three methods based on $ 10 $ runs for the application example.}
	\begin{center}
		\begin{tabular}{|l| c| c | c | c|} \hline
			&  BayesLCA & EM & SQP & Projected QN \\ \hline
			Log-Likelihood &  $ -781.8063 $   & $ -745.7291  $  &  $ -744.9672 $  & $ -746.8557 $   \\ \hline
			Number of Iterations & N.A. & $ 302 $ & $ 44 $   & $ 50 $ \\ \hline		
		\end{tabular}
	\end{center}
	\label{table:app_results}
\end{table}

From Table~\oldref{table:app_results}, SQP has the best performance in terms both the log-likelihood value and the number of iterations. The results from EM and Projected Quasi-Newton are very similar although EM needs way more iterations to converge. This agrees with the previous observations. We also note that all the three methods considered have larger log-likelihood values than that of BayesLCA. The method proposed by \cite{white2014bayeslca} actually has the smallest log-likelihood value.\\

In addition, since we do not know the true values, we computed pairwise root mean squared error (RMSE) based on the estimates, i.e, we compute RMSE of estimates for every two methods. Since we have considered four different methods, we will have six RMSEs, one number for each pair of methods. The results are reported in Table~\oldref{table:app_RMSE}

\begin{table}[htbp!]
	\caption{Pairwise root mean squared error (RMSE) of the four methods considered for the application example.}
	\begin{center}
		\begin{tabular}{|c| c| c | c | c|} \hline
			&  BayesLCA & EM & SQP & Projected QN \\ \hline
			BayesLCA &  $ 0 $   & $ 0.237  $  &  $ 0.241 $  & $ 0.233 $   \\ \hline
			EM & $0.237$ & $ 0 $ & $ 0.029 $   & $ 0.046 $ \\ \hline	
			SQP & $0.241$ & $ 0.029 $ & $ 0 $   & $ 0.045 $ \\ \hline	
			Projected QN & $0.233 $ & $ 0.046 $ & $ 0.045 $   & $ 0 $ \\ \hline		
		\end{tabular}
	\end{center}
	\label{table:app_RMSE}
\end{table}

The results in Table~\oldref{table:app_RMSE} are consistent with the observations we have made: Since the log-likelihood values are closer for EM, SQP and Projected QN, the pairwise RMSE of these three methods are way lower than those when paired with BayesLCA, for example the RMSE of SQP and EM is $0.029$, while the RMSE for SQP and BayesLCA is $0.241$, which is over eight times larger.

\section{Discussion}
In the previous section, we have shown the number of iterations of the proposed methods is smaller than that of the EM algorithm. In this section, we report the comparison of CPU times. Taking the application as an example, the runtime are reported in Table~\oldref{table:CPU}.

\begin{table}[htbp!]
	\caption{Comparison of CPU time (in seconds).}
	\begin{center}
		\begin{tabular}{|l| c| c | c |} \hline
			& EM & SQP & Projected QN \\ \hline
			CPU Time per Iteration & 0.08   & 0.31 & 0.39 \\ \hline
			Number of Iterations &   302    & 44   & 50      \\ \hline
			Overall Runtime & 24.2      & 13.6    & 19.5 \\ \hline
		\end{tabular}
	\end{center}
	\label{table:CPU}
\end{table}

From Table~\oldref{table:CPU}, we can see that the EM algorithm indeed has the lowest CPU time per iteration. However, when taking the number of iterations into account, the story is different: using SQP as example, the number of iterations of EM and SQP are $302$ and $ 44 $, respectively. The number of iterations for the SQP algorithm is around $ 1/8 $ of the EM algorithm, although the CPU time per iteration is merely about four times longer. Therefore, the total computational time of the SQP algorithm is significantly less than that of the EM algorithm. In the application example, the computational times of the SQP and Projected QN methods are respectively $43\%$ and $19\%$ better compared to the EM algorithm.

\section{Concluding Remarks} \label{sec:com}
The primary research objective of the paper is to provide alternative methods to learn the unknown parameters of the latent class model. Given the log-likelihood as a function of the parameters, we aim to find estimators that can maximize the log-likelihood function. The traditional way is to use the EM algorithm. However, it is observed that the EM algorithm converges slowly. Therefore, in this paper, we propose the use of two constrained optimization methods, namely the Sequential Quadratic Programming and the Projected Quasi-Newton methods as alternatives. Simulation studies and the real example in Section~\oldref{sec:sim} reveal that the two proposed methods perform well. The obvious advantages we observed are as follows:  (1) the two optimization methods produced slightly larger log-likelihood values compared to the EM algorithm; (2) they converge in significantly less iterations than the EM algorithm. That being said, we want to make it clear that the aim is not to completely replace the EM algorithm, rather we would like to provide alternative ways of achieving the same goal using some optimization methods. Inter-disciplinary collaboration between researchers in statistics and mathematical optimization has never been as important as in the big data era.

\bibliographystyle{elsarticle-harv}
\bibliography{sample}

\section*{About the Authors}
Hao Chen received his Ph.D. in Statistics from the University of British Columbia and is currently a~Senior~Data~Scientist at Precima. Lanshan Han holds a Ph.D. in Decision Sciences and Engineering Systems from the Rensselaer Polytechnic Institute and is currently a Director of Research and Development at Precima. Alvin Lim received his Ph.D. in Mathematical Sciences from the Johns Hopkins University and is currently Precima’s Chief Scientist and Vice President for Research and Development.

\newpage
\section*{Appendix}
The Python source codes for EM and Projected Quasi-Newton for LCM are available upon request.~The implementation of SQP is available in Python SciPy package.\\

The true weights and parameters used in Section~\oldref{sec:sim} are given below.

\subsection*{Example Bundle~1}

\begin{table}[H]
	\caption{ True Weights and Categorical Parameters for Example Bundle 1, $ d=1, K=2$.}
	\begin{center}
		\begin{tabular}{| c | c | c  c |  } \hline
			&  weights  & \multicolumn{2}{c|}{$d = 1$}  \\ \hline
			&           &  $0$ & $1$ \\ \hline
			$K=1$ &    $ 0.5$ & $0.4$ & $0.6$  \\
			$K=2$ &     $0.5$ & $0.8$ & $0.2$ \\
			\hline
		\end{tabular}
	\end{center}
	\label{table:b1:1}
\end{table}

\begin{table}[H]
	\caption{True Weights and Categorical Parameters for Example Bundle 1, $ d=1, K=3$.}
	\begin{center}
		\begin{tabular}{| c | c | c  c |  } \hline
			&  weights  & \multicolumn{2}{c|}{$d = 1$}  \\ \hline
			&           &  $0$ & $1$ \\ \hline
			$K=1$ &    $ 0.5$ & $0.4$ & $0.6$  \\
			$K=2$ &     $0.3$ & $0.8$ & $0.2$ \\
			$K=3$ &     $0.2$ & $0.1$ & $0.9$ \\
			\hline
		\end{tabular}
	\end{center}
	\label{table:b1:2}
\end{table}

\begin{table}[H]
	\caption{True Weights and Categorical Parameters for Example Bundle 1, $ d=2, K=2$.}
	\begin{center}
		\begin{tabular}{| c | c | c  c | c c|  } \hline
			&  weights  & \multicolumn{2}{c|}{$d = 1$} & \multicolumn{2}{c|}{$d = 2$} \\ \hline
			&           &  $0$ & $1$ & $0$ & $1$   \\ \hline
			$K=1$ &    $ 0.5$ & $0.4$ & $0.6$ & $0.1$ & $0.9$  \\
			$K=2$ &     $0.5$ & $0.8$ & $0.2$ & $0.6$ & $0.4$  \\
			\hline
		\end{tabular}
	\end{center}
	\label{table:b1:3}
\end{table}

\begin{table}[H]
	\caption{ True Weights and Categorical Parameters for Example Bundle 1, $ d=4, K=2$.}
	\begin{center}
		\begin{tabular}{| c | c | c  c | c c| c c| c c|  } \hline
			&  weights  & \multicolumn{2}{c|}{$d = 1$} & \multicolumn{2}{c|}{$d = 2$} & \multicolumn{2}{c|}{$d = 2$} & \multicolumn{2}{c|}{$d = 2$} \\ \hline
			&           &  $0$ & $1$ & $0$ & $1$ & $0$ & $1$ & $0$ & $1$  \\ \hline
			$K=1$ &    $ 0.5$ & $0.4$ & $0.6$ & $0.1$ & $0.9$ & $0.5$ & $0.5$ & $0.6$ & $0.4$  \\
			$K=2$ &     $0.5$ & $0.8$ & $0.2$ & $0.6$ & $0.4$ & $0.4$ & $0.6$ & $0.7$ & $0.3$ \\
			\hline
		\end{tabular}
	\end{center}
	\label{table:b1:4}
\end{table}

\subsection*{Example Bundle~2}

\begin{table}[H]
	\caption{True Weights and Categorical Parameters for Example Bundle 2, $ d=2, K=2$.}
	\begin{center}
		\begin{tabular}{| c | c | c  c | c c c| } \hline
			&  weights  & \multicolumn{2}{c|}{$d = 1$} & \multicolumn{3}{c|}{$d = 2$}  \\ \hline
			&           &  $0$ & $1$ &  $0$ & $1$ & $2$   \\ \hline
			$K=1$ &    $ 0.4$ & $0.1$ & $0.9$ & $0.8$ & $0.1$ & $0.1$   \\
			$K=2$ &     $0.6$ & $0.8$ & $0.2$ & $0.3$ & $0.4$ & $0.3$  \\
			\hline
		\end{tabular}
	\end{center}
	\label{table:b2:1}
\end{table}

\begin{table}[H]
	\caption{True Weights and Categorical Parameters for Example Bundle 2, $ d=2, K=3$.}
	\begin{center}
		\begin{tabular}{| c | c | c  c | c c c| } \hline
			&  weights  & \multicolumn{2}{c|}{$d = 1$} & \multicolumn{3}{c|}{$d = 2$}  \\ \hline
			&           &  $0$ & $1$ &  $0$ & $1$ & $2$   \\ \hline
			$K=1$ &    $ 0.4$ & $0.1$ & $0.9$ & $0.8$ & $0.1$ & $0.1$   \\
			$K=2$ &     $0.4$ & $0.8$ & $0.2$ & $0.3$ & $0.4$ & $0.3$  \\
			$K=3$ &     $0.2$ & $0.6$ & $0.4$ & $0.5$ & $0.3$ & $0.2$  \\
			\hline
		\end{tabular}
	\end{center}
	\label{table:b2:2}
\end{table}

\begin{table}[H]
	\caption{ True Weights and Categorical Parameters for Example Bundle 2, $ d=3, K=2$.}
	\begin{center}
		\begin{tabular}{| c | c | c  c | c c c| cc |  } \hline
			&  weights  & \multicolumn{2}{c|}{$d = 1$} & \multicolumn{3}{c|}{$d = 2$} & \multicolumn{2}{c|}{$d = 3$} \\ \hline
			&           &  $0$ & $1$ &  $0$ & $1$ & $2$ & $0$ & $1$   \\ \hline
			$K=1$ &    $ 0.4$ & $0.1$ & $0.9$ & $0.8$ & $0.1$ & $0.1$ & $0.6$ & $0.4$   \\
			$K=2$ &     $0.6$ & $0.8$ & $0.2$ & $0.3$ & $0.4$ & $0.3$ & $0.9$ & $0.1$ \\
			\hline
		\end{tabular}
	\end{center}
	\label{table:b2:3}
\end{table}

\begin{table}[H]
	\caption{ True Weights and Categorical Parameters for Example Bundle 2, $ d=3, K=3$.}
	\begin{center}
		\begin{tabular}{| c | c | c  c | c c c| cc |  } \hline
			&  weights  & \multicolumn{2}{c|}{$d = 1$} & \multicolumn{3}{c|}{$d = 2$} & \multicolumn{2}{c|}{$d = 3$} \\ \hline
			&           &  $0$ & $1$ &  $0$ & $1$ & $2$ & $0$ & $1$   \\ \hline
			$K=1$ &    $ 0.4$ & $0.1$ & $0.9$ & $0.8$ & $0.1$ & $0.1$ & $0.6$ & $0.4$   \\
			$K=2$ &     $0.4$ & $0.8$ & $0.2$ & $0.3$ & $0.4$ & $0.3$ & $0.9$ & $0.1$ \\
			$K=3$ &     $0.2$ & $0.6$ & $0.4$ & $0.6$ & $0.3$ & $0.1$ & $0.2$ & $0.8$ \\
			\hline
		\end{tabular}
	\end{center}
	\label{table:b2:4}
\end{table}

\subsection*{Example Bundle~3}

\begin{table}[H]
	\caption{ True Weights and Categorical Parameters for Example Bundle 3, $ d=3, K=3$.}
	\begin{center}
		\begin{tabular}{| c | c | c  c | c c | cc |  } \hline
			&  weights  & \multicolumn{2}{c|}{$d = 1$} & \multicolumn{2}{c|}{$d = 2$} & \multicolumn{2}{c|}{$d = 3$} \\ \hline
			&           &  $0$ & $1$ &  $0$ & $1$ & $0$ & $1$    \\ \hline
			$K=1$ &    $ 0.3$ & $0.9$ & $0.1$ & $0.3$ & $0.7$ & $0.1$ & $0.9$    \\
			$K=2$ &     $0.4$ & $0.2$ & $0.8$ & $0.5$ & $0.5$ & $0.55$ & $0.45$  \\
			$K=3$ &     $0.3$ & $0.1$ & $0.9$ & $0.4$ & $0.6$ & $0.3$ & $0.7$  \\
			\hline
		\end{tabular}
	\end{center}
	\label{table:b3:1}
\end{table}

\begin{table}[H]
	\caption{ True Weights and Categorical Parameters for Example Bundle 3, $ d=3, K=4$.}
	\begin{center}
		\begin{tabular}{| c | c | c  c | c c | cc |  } \hline
			&  weights  & \multicolumn{2}{c|}{$d = 1$} & \multicolumn{2}{c|}{$d = 2$} & \multicolumn{2}{c|}{$d = 3$} \\ \hline
			&           &  $0$ & $1$ &  $0$ & $1$ & $0$ & $1$    \\ \hline
			$K=1$ &    $ 0.3$ & $0.9$ & $0.1$ & $0.3$ & $0.7$ & $0.1$ & $0.9$    \\
			$K=2$ &     $0.2$ & $0.2$ & $0.8$ & $0.5$ & $0.5$ & $0.55$ & $0.45$  \\
			$K=3$ &     $0.3$ & $0.1$ & $0.9$ & $0.4$ & $0.6$ & $0.3$ & $0.7$  \\
			$K=4$ &     $0.2$ & $0.5$ & $0.5$ & $0.9$ & $0.1$ & $0.2$ & $0.8$  \\
			\hline
		\end{tabular}
	\end{center}
	\label{table:b3:2}
\end{table}

\begin{table}[H]
	\caption{ True Weights and Categorical Parameters for Example Bundle 3, $ d=4, K=4$.}
	\begin{center}
		\begin{tabular}{| c | c | c  c | c c | cc | cc|  } \hline
			&  weights  & \multicolumn{2}{c|}{$d = 1$} & \multicolumn{2}{c|}{$d = 2$} & \multicolumn{2}{c|}{$d = 3$} & \multicolumn{2}{c|}{$d = 4$} \\ \hline
			&           &  $0$ & $1$ &  $0$ & $1$ & $0$ & $1$ &  $0$ & $1$     \\ \hline
			$K=1$ &    $ 0.3$ & $0.9$ & $0.1$ & $0.3$ & $0.7$ & $0.1$ & $0.9$ & $0.6$ & $0.4$   \\
			$K=2$ &     $0.2$ & $0.2$ & $0.8$ & $0.5$ & $0.5$ & $0.55$ & $0.45$ & $0.5$ & $0.5$  \\
			$K=3$ &     $0.3$ & $0.1$ & $0.9$ & $0.4$ & $0.6$ & $0.3$ & $0.7$ & $0.7$ & $0.3$ \\
			$K=4$ &     $0.2$ & $0.5$ & $0.5$ & $0.9$ & $0.1$ & $0.2$ & $0.8$ & $0.5$ & $0.5$  \\
			\hline
		\end{tabular}
	\end{center}
	\label{table:b3:3}
\end{table}

\begin{table}[H]
	\caption{ True Weights and Categorical Parameters for Example Bundle 3, $ d=5, K=3$.}
	\begin{center}
		\begin{tabular}{| c | c | c  c | c c | cc | cc| cc|  } \hline
			&  weights  & \multicolumn{2}{c|}{$d = 1$} & \multicolumn{2}{c|}{$d = 2$} & \multicolumn{2}{c|}{$d = 3$} & \multicolumn{2}{c|}{$d = 4$} & \multicolumn{2}{c|}{$d = 5$} \\ \hline
			&           &  $0$ & $1$ &  $0$ & $1$ & $0$ & $1$ &  $0$ & $1$ & $0$ & $1$    \\ \hline
			$K=1$ &    $ 0.3$ & $0.9$ & $0.1$ & $0.3$ & $0.7$ & $0.1$ & $0.9$ & $0.6$ & $0.4$ & $0.7$ & $0.3$   \\
			$K=2$ &     $0.4$ & $0.2$ & $0.8$ & $0.5$ & $0.5$ & $0.55$ & $0.45$ & $0.5$ & $0.5$ & $0.3$ & $0.7$  \\
			$K=3$ &     $0.3$ & $0.1$ & $0.9$ & $0.4$ & $0.6$ & $0.3$ & $0.7$ & $0.9$ & $0.1$ & $0.2$ & $0.8$ \\
			\hline
		\end{tabular}
	\end{center}
	\label{table:b3:4}
\end{table}

\subsection*{Example Bundle~4}

\begin{table}[H]
	\caption{ True Weights and Categorical Parameters for Example Bundle 4, $ d=4, K=4$.}
	\begin{center}
		\begin{tabular}{| c | c | c  c | c c | cc | cc|  } \hline
			&  weights  & \multicolumn{2}{c|}{$d = 1$} & \multicolumn{2}{c|}{$d = 2$} & \multicolumn{2}{c|}{$d = 3$} & \multicolumn{2}{c|}{$d = 4$} \\ \hline
			&           &  $0$ & $1$ &  $0$ & $1$ & $0$ & $1$ &  $0$ & $1$     \\ \hline
			$K=1$ &    $ 0.3$ & $0.9$ & $0.1$ & $0.3$ & $0.7$ & $0.1$ & $0.9$ & $0.6$ & $0.4$   \\
			$K=2$ &     $0.2$ & $0.2$ & $0.8$ & $0.5$ & $0.5$ & $0.55$ & $0.45$ & $0.5$ & $0.5$  \\
			$K=3$ &     $0.3$ & $0.1$ & $0.9$ & $0.4$ & $0.6$ & $0.3$ & $0.7$ & $0.7$ & $0.3$ \\
			$K=4$ &     $0.2$ & $0.5$ & $0.5$ & $0.9$ & $0.1$ & $0.2$ & $0.8$ & $0.5$ & $0.5$  \\
			\hline
		\end{tabular}
	\end{center}
	\label{table:b4:1}
\end{table}

\begin{table}[H]
	\caption{ True Weights and Categorical Parameters for Example Bundle 4, $ d=4, K=5$.}
	\begin{center}
		\begin{tabular}{| c | c | c  c | c c | cc | cc|  } \hline
			&  weights  & \multicolumn{2}{c|}{$d = 1$} & \multicolumn{2}{c|}{$d = 2$} & \multicolumn{2}{c|}{$d = 3$} & \multicolumn{2}{c|}{$d = 4$} \\ \hline
			&           &  $0$ & $1$ &  $0$ & $1$ & $0$ & $1$ &  $0$ & $1$     \\ \hline
			$K=1$ &    $ 0.3$ & $0.9$ & $0.1$ & $0.3$ & $0.7$ & $0.1$ & $0.9$ & $0.6$ & $0.4$   \\
			$K=2$ &     $0.2$ & $0.2$ & $0.8$ & $0.5$ & $0.5$ & $0.55$ & $0.45$ & $0.5$ & $0.5$  \\
			$K=3$ &     $0.3$ & $0.1$ & $0.9$ & $0.4$ & $0.6$ & $0.3$ & $0.7$ & $0.7$ & $0.3$ \\
			$K=4$ &     $0.1$ & $0.5$ & $0.5$ & $0.9$ & $0.1$ & $0.2$ & $0.8$ & $0.5$ & $0.5$  \\
			$K=5$ &     $0.1$ & $0.8$ & $0.2$ & $0.1$ & $0.9$ & $0.9$ & $0.1$ & $0.7$ & $0.3$  \\
			\hline
		\end{tabular}
	\end{center}
	\label{table:b4:2}
\end{table}

\begin{table}[H]
	\caption{ True Weights and Categorical Parameters for Example Bundle 4, $ d=5, K=4$.}
	\begin{center}
		\begin{tabular}{| c | c | c  c | c c | cc | cc| cc|  } \hline
			&  weights  & \multicolumn{2}{c|}{$d = 1$} & \multicolumn{2}{c|}{$d = 2$} & \multicolumn{2}{c|}{$d = 3$} & \multicolumn{2}{c|}{$d = 4$} & \multicolumn{2}{c|}{$d = 5$} \\ \hline
			&           &  $0$ & $1$ &  $0$ & $1$ & $0$ & $1$ &  $0$ & $1$ & $0$& $1$      \\ \hline
			$K=1$ &    $ 0.3$ & $0.9$ & $0.1$ & $0.3$ & $0.7$ & $0.1$ & $0.9$ & $0.6$ & $0.4$ & $0.2$ & $0.8$    \\
			$K=2$ &     $0.2$ & $0.2$ & $0.8$ & $0.5$ & $0.5$ & $0.55$ & $0.45$ & $0.5$ & $0.5$ & $0.8$ & $0.2$  \\
			$K=3$ &     $0.3$ & $0.1$ & $0.9$ & $0.4$ & $0.6$ & $0.3$ & $0.7$ & $0.7$ & $0.3$ & $0.3$ & $0.7$ \\
			$K=4$ &     $0.2$ & $0.5$ & $0.5$ & $0.9$ & $0.1$ & $0.2$ & $0.8$ & $0.5$ & $0.5$ & $0.9$ & $0.1$ \\
			\hline
		\end{tabular}
	\end{center}
	\label{table:b4:3}
\end{table}

\begin{table}[H]
	\caption{ True Weights and Categorical Parameters for Example Bundle 4, $ d=5, K=5$.}
	\begin{center}
		\begin{tabular}{| c | c | c  c | c c | cc | cc| cc|  } \hline
			&  weights  & \multicolumn{2}{c|}{$d = 1$} & \multicolumn{2}{c|}{$d = 2$} & \multicolumn{2}{c|}{$d = 3$} & \multicolumn{2}{c|}{$d = 4$} & \multicolumn{2}{c|}{$d = 5$} \\ \hline
			&           &  $0$ & $1$ &  $0$ & $1$ & $0$ & $1$ &  $0$ & $1$ & $0$& $1$      \\ \hline
			$K=1$ &    $ 0.3$ & $0.9$ & $0.1$ & $0.3$ & $0.7$ & $0.1$ & $0.9$ & $0.6$ & $0.4$ & $0.4$ & $0.6$    \\
			$K=2$ &     $0.2$ & $0.2$ & $0.8$ & $0.5$ & $0.5$ & $0.55$ & $0.45$ & $0.5$ & $0.5$ & $0.7$ & $0.3$  \\
			$K=3$ &     $0.3$ & $0.1$ & $0.9$ & $0.4$ & $0.6$ & $0.3$ & $0.7$ & $0.7$ & $0.3$ & $0.4$ & $0.6$ \\
			$K=4$ &     $0.1$ & $0.5$ & $0.5$ & $0.9$ & $0.1$ & $0.2$ & $0.8$ & $0.5$ & $0.5$ & $0.8$ & $0.2$ \\
			$K=5$ &     $0.1$ & $0.8$ & $0.2$ & $0.1$ & $0.9$ & $0.9$ & $0.1$ & $0.7$ & $0.3$ & $0.9$  & $0.1$ \\
			\hline
		\end{tabular}
	\end{center}
	\label{table:b4:4}
\end{table}

\end{document}